\documentclass{article}

\usepackage[preprint]{neurips_2026}

\usepackage[utf8]{inputenc}
\usepackage[T1]{fontenc}
\usepackage{hyperref}
\usepackage{url}
\usepackage{booktabs}
\usepackage{amsfonts}
\usepackage{amsmath}
\usepackage{nicefrac}
\usepackage{microtype}
\usepackage{xcolor}
\usepackage{graphicx}
\usepackage{algorithm}
\usepackage{algorithmic}
\usepackage{tikz}
\usepackage{pgfplots}
\pgfplotsset{compat=1.18}
\usetikzlibrary{arrows.meta, positioning, shapes.geometric, backgrounds, fit}
\usepackage{enumitem}
\usepackage{colortbl}
\usepackage{multirow}
\usepackage{amsthm}

\title{Trace2Policy: From Expert Behavior Traces to\\Self-Evolving Decision Agents}

\author{
  Junli Zha \quad Jinbo Wang \quad Chao Zhou \quad Xiang Song \\[2pt]
  SF Express \\[2pt]
  \texttt{\{zhajunli, wangjinbo, charleszhou, songxiang1\}@sf-express.com}
}

\begin{document}
\maketitle

\begin{abstract}
Decision rules that enterprise experts apply tacitly---in auditing, compliance, and contract review---can be systematically recovered and improved through iterative error analysis. We present \textbf{Trace2Policy}, whose core mechanism---\textbf{EISR} (\textbf{E}rror-driven \textbf{I}terative \textbf{S}kill \textbf{R}efinement)---maintains a human-readable rule document as its optimization target: each round executes the rules on a validation set, clusters errors by root cause into MISSING, WRONG, or CONFLICT types, applies targeted patches, and commits only those that pass a regression gate. \textbf{For this class of compliance-sensitive, skewed-base-rate decision tasks, we identify rule quality---not model capability---as the dominant performance lever}: across five LLMs, one-shot distillation plateaus near $\sim$70\% on the deployed pool, while eight EISR rounds lift the same rules to 79.6\% when compiled into deterministic Python---zero LLM calls at inference. \textbf{Execution form compounds the gain: in production, the same EISR-refined content runs 9.8~pp higher as compiled Python than as an LLM prompt, a form-and-engineering bundle the 22-day deployment matured together.} Deployed for 22 days at a major logistics carrier (3,349 audit cases), the compiled pipeline outperforms the pure-LLM baseline it replaced (72.7\%); on these calibrated, skewed-base-rate workloads, re-enabling LLM fallback monotonically degrades accuracy. An LLM-driven variant, \textbf{Auto-EISR}, reproduces this refinement at \$5--\$10 per cycle versus $\sim$70 expert-hours, and transfers to four public benchmarks spanning legal reasoning (LegalBench) and process-mining decisions (BPIC 2012) without re-engineering.
\end{abstract}

\section{Introduction}
\label{sec:intro}

A vast category of enterprise work consists of \emph{judgment-intensive decision tasks}: an expert navigates multiple information systems, cross-references data fields with domain-specific semantics, and applies decision heuristics accumulated over years of practice.
Examples include damage liability auditing in logistics, insurance claim adjudication, regulatory compliance review, and manufacturing quality inspection.
These tasks share three characteristics: (1) they follow implicit but systematic rules, (2) the rules interact with specific software systems whose behavior contains hidden conventions, and (3) the rules evolve as business conditions change.

Automating such tasks requires solving three problems that existing approaches address only partially:

\textbf{Knowledge acquisition: learning \emph{what to decide}, not just \emph{what to click}.}
GUI agents such as CogAgent~\citep{hong2024cogagent}, SeeAct~\citep{zheng2024seeact}, UI-TARS~\citep{qin2025uitars}, and DigiRL~\citep{bai2024digirl} learn to operate interfaces---which buttons to click, which fields to fill---but not the decision logic behind those operations.
AgentTrek~\citep{agenttrek2025} synthesizes trajectories from web tutorials, but tutorials describe procedures, not judgment heuristics.
Process Mining~\citep{van2016process} discovers activity flows from event logs but not why particular decisions were made.
Skill library approaches like Voyager~\citep{wang2023voyager} accumulate reusable code skills in game environments, but enterprise decisions require interpretable rules, not executable programs.
We need to extract \emph{decision policies}---the why behind expert actions---from natural work behavior.

\textbf{Knowledge refinement: why one-shot extraction fails.}
Self-Refine~\citep{madaan2023selfrefine} and Reflexion~\citep{shinn2023reflexion} iteratively improve LLM \emph{outputs}; TextGrad~\citep{yehudai2025textgrad} and OPRO~\citep{yang2024opro} optimize \emph{prompts}; AgentRefine~\citep{pan2025agentrefine} refines agent \emph{behavior} through environment feedback.
But none refine \emph{externalized, interpretable decision rules}---they improve model performance without producing auditable knowledge artifacts.
One-shot extraction---whether through LLM distillation, few-shot learning, or expert interviews---captures only \emph{surface knowledge}: the obvious procedures and common patterns.
Our experiments show this surface knowledge is insufficient: on the 111-waybill validation pool, v1 Skills $\approx$ no rules $\approx$ few-shot, all $\sim$70\% across 5 models, and on a 139-case held-out, unrefined rules can actively harm strong models through ``authority displacement'' (Section~\ref{sec:exp}).
\emph{Deep knowledge}---system encoding conventions, implicit action semantics, claim-action mismatches---requires iterative error analysis to surface as auditable artifacts.

\textbf{Knowledge evolution: beyond static deployment.}
Recent surveys on self-evolving agents~\citep{tao2025selfevolving} identify a critical gap: most agent systems remain static after deployment, unable to adapt to changing business conditions.
Existing evolution mechanisms (RL fine-tuning, meta-learning, population-based methods) require costly labeled feedback.
Enterprise decision tasks offer a unique opportunity: human reviewers already examine every case as part of their existing workflow, providing \emph{natural ground truth at zero marginal cost}.
We need a framework that exploits this signal for continuous self-improvement.

\begin{figure}[t]
\centering
\resizebox{\columnwidth}{!}{%
\begin{tikzpicture}[
    node distance=0.4cm and 0.6cm,
    phase/.style={rectangle, rounded corners=4pt, draw=#1, fill=#1!8, minimum height=1.4cm, minimum width=2.2cm, align=center, font=\small\bfseries, line width=0.8pt},
    subphase/.style={rectangle, rounded corners=3pt, draw=#1, fill=#1!5, minimum height=0.8cm, minimum width=2.0cm, align=center, font=\scriptsize, line width=0.6pt, dashed},
    arr/.style={-{Stealth[length=5pt]}, thick, color=#1},
    lbl/.style={font=\scriptsize, align=center, text width=2.0cm},
]
\node[phase=blue!70!black] (p0) {Phase 0\\[-1pt]{\footnotesize Trace}};
\node[phase=teal!70!black, right=0.8cm of p0] (p1) {Phase 1\\[-1pt]{\footnotesize Structure}};
\node[phase=orange!70!black, right=0.8cm of p1] (p2) {Phase 2\\[-1pt]{\footnotesize Distill}};
\node[phase=red!70!black, right=0.8cm of p2] (p3) {Phase 3\\[-1pt]{\footnotesize Refine}};
\node[phase=green!50!black, right=0.8cm of p3] (p4) {Phase 4\\[-1pt]{\footnotesize Evolve}};

\node[subphase=red!55!black, above=0.5cm of p3, xshift=-1.3cm] (p3h) {Human-EISR\\\S\ref{sec:exp}};
\node[subphase=purple!55!black, above=0.5cm of p3, xshift=1.3cm] (p3a) {Auto-EISR\\\S\ref{sec:auto-eisr}};
\draw[arr=red!40!black, dashed, line width=0.4pt] (p3.north) -- (p3h.south);
\draw[arr=purple!40!black, dashed, line width=0.4pt] (p3.north) -- (p3a.south);

\draw[arr=gray!60!black] (p0) -- (p1);
\draw[arr=gray!60!black] (p1) -- (p2);
\draw[arr=gray!60!black] (p2) -- (p3);
\draw[arr=gray!60!black] (p3) -- (p4);

\node[lbl, below=0.15cm of p0] {Agent Observer\\multimodal\\behavior capture};
\node[lbl, below=0.15cm of p1] {VLM\\traces $\to$\\decision records};
\node[lbl, below=0.15cm of p2] {LLM\\records $\to$\\initial policy};
\node[lbl, below=0.15cm of p3] {Error-driven\\iterative\\refinement};
\node[lbl, below=0.15cm of p4] {Deploy +\\natural data\\flywheel};

\draw[arr=green!50!black, dashed, line width=1pt] (p4.south) -- ++(0,-1.7) -| node[below, pos=0.25, font=\scriptsize\itshape, text=green!40!black]{Continuous Self-Evolution} (p3.south);
\end{tikzpicture}
}
\caption{The Trace2Policy pipeline. Expert behavior traces are progressively transformed into executable, self-evolving decision policies. Phase 3 admits two implementations---a human-in-the-loop refiner (\S\ref{sec:exp}, as originally conceived) and an LLM-driven Auto-EISR variant (\S\ref{sec:auto-eisr}); both feed the same Phase 4 flywheel, which in turn feeds back into Phase 3 for continuous improvement.}
\label{fig:framework}
\end{figure}
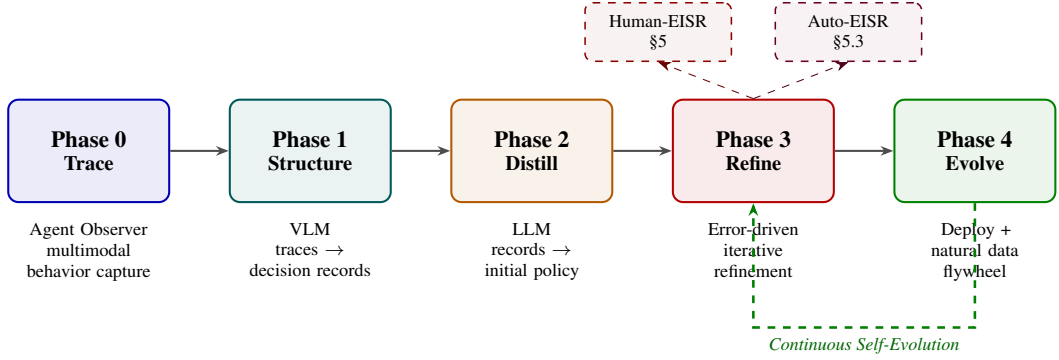

We present \textbf{Trace2Policy}, an end-to-end framework anchored on \textbf{Error-driven Iterative Skill Refinement} (\textbf{EISR}): a structured diagnose-and-patch loop that treats an externalized rule document as the artifact to optimize, classifies each validation error into a MISSING, WRONG, or CONFLICT root-cause cluster, proposes a targeted patch per cluster, and accepts the patch only if a regression gate confirms it does not break previously-correct cases (Algorithm~\ref{alg:eisr}). Surrounding EISR are upstream components that supply its inputs---an Agent Observer that records expert behavior, a VLM that structurizes raw traces into decision records, and an LLM distiller that extracts an initial $v_1$ rule document from a handful of labeled examples---and a downstream production data flywheel in which ongoing human reviews provide natural ground truth at zero marginal cost. The full pipeline is visualized in Figure~\ref{fig:framework}; component details are in \S\ref{sec:method}. We further show (\S\ref{sec:compilation-bridge}) that EISR's output admits two complementary execution forms---an LLM prompt and compiled deterministic Python---and report a within-deployment form-bundle observation: a $9.8$\,pp accuracy gap that bundles form with engineering and is scoped to the production regime.

Our contributions:
\begin{itemize}[nosep]
    \item \textbf{Trace2Policy}: an end-to-end pipeline from raw expert behavior traces to deployed decision agents, validated through a 22-day production deployment covering 3,349 resolved cases at a major logistics carrier---to our knowledge, the first such study at this scale in the dialogue policy literature.
    \item \textbf{EISR (Error-driven Iterative Skill Refinement)}: a rule refinement algorithm using LLM-as-optimizer with regression-gated acceptance. We provide preliminary evidence of its effectiveness: Auto-EISR matches Human-EISR on action and category accuracy at \$5--\$10 per refinement cycle vs $\sim$70 expert-hours.
    \item \textbf{Within-regime observations on rule quality and execution form}: across five model scales, the variance attributable to rule version exceeds that attributable to model choice (Table~\ref{tab:baseline}); within the production regime, compiled-Python execution with 22-day accumulated extensions yields a $9.8$\,pp gap over the skill-level prompt (Table~\ref{tab:prod-evolution}). We treat this as a form-bundle observation rather than a clean causal claim; isolating the form-only component is a measurement gap, and cross-regime behavior is reported in \S\ref{sec:compilation-bridge}.
    \item \textbf{An authority displacement observation} and documentation of three categories of ``trap rules''---deep knowledge invisible to one-shot extraction---with implications for human-AI collaboration design.
\end{itemize}

\textbf{Contribution type.} We submit under the \emph{Use-Inspired} type: the use case arises from pre-existing operational needs, methodological choices are designed against properties of this use case, and our evaluation includes both ML baselines and the human-expert workflow the system augments.

\textbf{Scope.} The abstract's ``binding constraint'' phrasing scopes to the within-deployment regime: \S\ref{sec:compilation-bridge} reports that the compiled pipeline reaches $74.1\%$ on the $139$-case skill-level held-out (below B1b $82.7\%$), which we treat as the explicit \emph{scope of the claim}, not a contradiction. The $9.8$\,pp ``research lever'' bundles form with engineering, and isolating the form-only component is flagged as a measurement gap. Cross-domain transferability is probed on four public benchmarks (\S\ref{sec:cross-domain}); rigorous multi-domain validation with domain-specific baselines remains open.

\section{Related Work}
\label{sec:related}

\textbf{Operating interfaces vs.\ extracting decisions.}
GUI agents~\citep{hong2024cogagent,zheng2024seeact,qin2025uitars,bai2024digirl} and trajectory synthesizers (AgentTrek~\citep{agenttrek2025}) learn \emph{how to click}; Voyager~\citep{wang2023voyager} accumulates code skills in games. Process Mining~\citep{van2016process} discovers \emph{what} was done from event logs but not \emph{why}. None extract interpretable decision rules from natural expert behavior in production enterprise systems---which is what Trace2Policy targets.

\textbf{What gets refined.}
Self-Refine~\citep{madaan2023selfrefine}, Reflexion~\citep{shinn2023reflexion} refine LLM \emph{outputs}; TextGrad~\citep{yehudai2025textgrad}, OPRO~\citep{yang2024opro} refine \emph{prompts}; AgentRefine~\citep{pan2025agentrefine} refines agent \emph{behavior}.
EISR refines \emph{externalized, version-controlled decision rules} that exist as auditable artifacts outside any particular model---refinements are human-inspectable, the same rules apply across 6 LLMs we evaluate, and the knowledge persists across model deployments.
The pre-flight validator and regression gate borrow from AI Scientist v2~\citep{sakana2025aiscientistv2}, FunSearch~\citep{funsearch2024}, and Retroformer's~\citep{retroformer2024} retrospective/actor split.

\textbf{Prompt optimization.}
DSPy~\citep{khattab2023dspy} and MIPRO~\citep{opsahl2024mipro} optimize prompt templates via metric-driven search. EISR differs in two key respects: (1)~\emph{Artifact type}---DSPy optimizes parametric prompt chains, whereas EISR refines an externalized, human-readable rule document amenable to compliance audit and version control; (2)~\emph{Optimization objective}---DSPy maximizes end-to-end task metrics through prompt tuning, whereas EISR optimizes rule coverage and precision under interpretability constraints. A systematic comparison would require constructing equivalent evaluation protocols across these fundamentally different artifact types, which we leave to future work.

\textbf{Self-evolving agents.}
The 2025 survey~\citep{tao2025selfevolving} organizes existing systems along axes (parameters, prompts, memory, tools, workflows); most evolve model internals and require explicit reward or human feedback.
We evolve \emph{externalized decision rules} via \emph{natural ground truth} from existing human review workflows.

\textbf{Concurrent work (2025--2026).}
Jiang et al.~\citep{jiang2026sok} systematize the agentic skill lifecycle; Jiang et al.~\citep{jiang2025adaptation} survey post-training adaptation; Zhang et al.~\citep{zhang2026experience} identify a ``missing diagonal'' for declarative-rule extraction; Xu et al.~\citep{xu2026agentskills}, Bi et al.~\citep{bi2026skillmining} survey/mine skill artifacts; Nian et al.~\citep{nian2026auditable} formalize agent-auditability. We provide a deployed instantiation with production data. Positioning summary in Appendix~\ref{app:positioning}.

\section{The Trace2Policy Framework}
\label{sec:method}

\subsection{Phase 0--1: Behavior Capture and Structurization}

An \emph{Agent Observer} system passively captures expert work as multimodal event streams (window focus, mouse clicks with region screenshots, keyboard input, browser events), segmented into task-level trajectories using business anchors (e.g., waybill numbers).
In our case study, this yields 555 trajectories from two expert auditors performing the same task (Appendix~\ref{app:phases01}).

A VLM then converts raw trajectories into structured decision records---systems consulted, key observations, reasoning chains, and evidence.
This yields 476 usable records (85.8\% success rate).
The 476 structured records feed policy distillation (Phase 2), while the full 555 records provide ground truth for evaluation using compact API data (Appendix~\ref{app:input-repr}).

\subsection{Phase 2: Automated Policy Distillation}

From $N$ structured decision records, an LLM distills an initial policy---an executable rule base we call \emph{Skills} (the document) containing individual \emph{rules} (the numbered bullet points inside).
Throughout the paper we count rules at the bullet-point granularity: the v8 policy contains 62 rules distributed across 5 Markdown files (one per decision path plus one for path routing).
Skills are organized in three layers:

\textbf{Workflow layer}: which systems to query, in what order, with what parameters.
For example: ``First query AuditPlatform\footnote{AuditPlatform, RouteSystem, and DetectorService are pseudonyms for the carrier's internal systems.} for liability determination info, then query RouteSystem for route remarks and image tags.''

\textbf{Decision layer}: path routing rules and judgment conditions.
Path routing determines \emph{which} decision logic to apply (e.g., ``if feedback mentions intact outer packaging $\wedge$ original code = \textsc{PackagingDamage}, route to Path A: packaging judgment error'').
Judgment conditions within each path determine the outcome (approve/reject) and fine-grained category.

\textbf{Platform layer}: API endpoints, request/response schemas, field encoding mappings (e.g., \textsc{PackagingDamage} and \textsc{SharedLiability} are anonymized labels for two specific liability codes that figure prominently in the trap-rule analysis below), and login procedures.

In our case study, Phase 2 produces 62 judgment rules across 4 decision paths (E: special processes, C: reporting/compliance, A: packaging judgment errors, BD: general appeals), covering 41 judgment situations and 17 outcome codes.

\textbf{The surface knowledge trap.}
These 62 rules represent the \emph{surface layer}---patterns visible in individual records.
v1 Skills achieve only 69.2\% average action accuracy across 5 models, statistically indistinguishable from no rules (70.3\%).
However, ``insufficient'' is \emph{form-dependent}: the same surface knowledge, compiled into Python, yields $79.6 / 77.3 / 76.8\%$ on three production benchmarks with $0\%$ LLM calls (Section~\ref{sec:production}).
Refinement deepens rule content; compilation enforces it against the LLM's unrelated prior.

\subsection{Phase 3: EISR --- Error-driven Iterative Skill Refinement}
\label{sec:eisr}

EISR is the refinement engine that discovers deep knowledge by systematically analyzing prediction errors (Algorithm~\ref{alg:eisr}).

\begin{algorithm}[t]
\caption{EISR: Error-driven Iterative Skill Refinement}
\label{alg:eisr}
\begin{algorithmic}[1]
\REQUIRE Initial policy $S_0$, validation set $V$, ground truth $\text{GT}$, max rounds $K$
\ENSURE Refined policy $S^*$
\STATE $S \leftarrow S_0$
\FOR{$t = 1$ to $K$}
    \STATE $R \leftarrow \textsc{Execute}(S, V)$ \COMMENT{Agent applies policy to each case}
    \STATE $E \leftarrow \textsc{Compare}(R, \text{GT})$ \COMMENT{Identify errors}
    \IF{$|E| = 0$}
        \STATE \textbf{break} \COMMENT{Converged}
    \ENDIF
    \STATE $D \leftarrow \textsc{Diagnose}(S, E)$ \COMMENT{Cluster errors by root cause}
    \STATE $S \leftarrow \textsc{Refine}(S, D)$ \COMMENT{Targeted fixes}
\ENDFOR
\RETURN $S$
\end{algorithmic}
\end{algorithm}

\textbf{Structured Error Diagnosis.}
The \textsc{Diagnose} step classifies each error as \textsc{Missing} (no rule covers this case), \textsc{Wrong} (a rule matches but prescribes the wrong action), or \textsc{Conflict} (multiple rules match with contradictory actions).
Crucially, similar errors are \emph{clustered by root cause} before fixing.
This clustering reduces the number of fix attempts from $|E|$ (one per error) to $m$ (one per root cause, where $m \ll |E|$), reducing regression risk proportionally (Appendix~\ref{sec:theory}).

\textbf{Trap rules discovered.}
Across 8 EISR rounds, three ``trap rules'' were discovered---deep knowledge invisible to any one-shot method:

\emph{Trap 1: Encoding State Ambiguity (R5$\to$R6).}
The AuditPlatform system displays \emph{current} liability codes, not original ones.
A liability code displayed as \textsc{SharedLiability} may have just been changed from \textsc{PackagingDamage} during the review.
This caused a systematic crash from 90\% to 70\% in R5; the fix (``judge by reviewer action words, not current codes'') restored accuracy in R6.

\emph{Trap 2: Implicit Action Prefix (R5$\to$R6).}
A ``C:'' prefix in reviewer comments is an undocumented signal that the reviewer is \emph{executing} a liability reassignment---an implicit approval hidden in text formatting.

\emph{Trap 3: Claim Mismatch (R7$\to$R8).}
Headquarters performing action B $\neq$ accepting appellant's request for action A.
``Doing something'' $\neq$ ``accepting the specific claim.''

These rules share a key property: they encode \emph{system behavior conventions} that experts internalize unconsciously but that no documentation describes and no amount of one-shot analysis can extract. Formal analysis of EISR as version-space narrowing and sample-complexity bounds are in Appendix~\ref{sec:theory}.

\subsection{Phase 4: Production Deployment and Self-Evolution}
\label{sec:flywheel}

The final phase deploys the refined policy and establishes a \emph{natural data flywheel} for continuous evolution.
The agent provides reference answers; human auditors still review every case as part of their unchanged workflow, generating ground truth at zero marginal annotation cost.
Errors are automatically clustered, and when similar errors accumulate ($\geq 5$ cases), an incremental EISR round triggers---typically a single iteration modifying 1--2 rules, validated via regression on historical cases.
As accuracy improves, high-confidence cases can gradually bypass human review, creating a virtuous cycle.
Full details and cold-start vs.\ incremental comparison are in Appendix~\ref{app:flywheel}.

\section{Production Pipeline Study: The Compilation Level}
\label{sec:production}

This section evaluates Trace2Policy's pipeline-level instantiation---EISR output compiled into $1{,}193$ lines of Python---on three real benchmarks over $22$ days of deployment.

\textbf{Setup.} The production system is a rule-first cascade applying $\sim$9 rule layers to each waybill's AuditPlatform data plus RouteSystem/image/detector signals, with an LLM fallback gate that is dormant in deployment ($0\%$ LLM calls; \S\ref{sec:prod-c} activates it). Rule layers were authored by a human curator over 22 days of daily EISR. We evaluate on three labeled sets: \texttt{benchmark\_1000} (993 waybills, the training benchmark), \texttt{new\_2000} (1{,}462, a held-out set constructed immediately before the 22-day window), and \texttt{recent\_week} (894, drift); details in Appendix~\ref{app:flywheel}.

\subsection{Production Evolution}
\label{sec:prod-evolution}

Table~\ref{tab:prod-evolution} summarizes $22$ days of EISR-driven rule refinement. A critical reading note: earlier iteration accuracies were recorded on different benchmarks as the team rotated through labeled sets. A naive concatenation of these numbers produced an illusory ``$81.1 \to 77.3\%$ regression'' which disappears once all versions are measured on the same benchmark.

\begin{table}[t]
\centering
\caption{Production pipeline vs.\ baselines. ``Acc'' = accuracy; ``Rej-F1'' = rejection-class F1 (the audit-relevant metric). Always-approve is a degenerate reference (zero rejections). Historical pipeline trajectory and full bootstrap CIs in Appendix~\ref{app:flywheel}.}
\label{tab:prod-evolution}
\small
\setlength{\tabcolsep}{4pt}
\begin{tabular}{lcccccccc}
\toprule
 & \multicolumn{2}{c}{\texttt{benchmark\_1000}} & \multicolumn{2}{c}{\texttt{new\_2000}} & \multicolumn{2}{c}{\texttt{recent\_week}} & LLM \\
\cmidrule(lr){2-3}\cmidrule(lr){4-5}\cmidrule(lr){6-7}
Configuration & Acc & Rej-F1 & Acc & Rej-F1 & Acc & Rej-F1 & calls \\
              & ($n{=}993$) & & ($n{=}1462$) & & ($n{=}894$) & & \\
\midrule
Always-approve (degenerate) & $74.6\%$ & $0.00$ & $76.6\%$ & $0.00$ & $74.6\%$ & $0.00$ & $0\%$ \\
Opus zero-shot (B1b)              & $43.8 {\pm} 0.5\%$ & $0.32$ & --- & --- & --- & --- & $100\%$ \\
Opus $+$ skills\_v8 prompt         & $69.8 {\pm} 0.1\%$ & $0.17$ & --- & --- & --- & --- & $100\%$ \\
Pre-deployment pure LLM            & $72.7\%$ & --- & --- & --- & --- & --- & $100\%$ \\
\textbf{Pipeline \texttt{current}} & $\mathbf{79.6\%}$ & $\mathbf{0.50}$ & $\mathbf{77.3\%}$ & $\mathbf{0.38}$ & $\mathbf{76.8\%}$ & $\mathbf{0.43}$ & $\mathbf{0\%}$ \\
\bottomrule
\end{tabular}

\vspace{0.2em}
\raggedright{\footnotesize$^{\dagger}$Opus baselines are 3 independent seeds at $T{=}0.3$; mean $\pm$ std. Mean scored rates: B1b $980/993$ ($13$ HTTP 429/parse failures); full-rules prompt $939/993$ ($54$ failures). Accuracy is over scored cases; under the most generous unscored-recovery assumption, upper bounds are $\sim$44.5\% and $\sim$71.4\% respectively. Pipeline accuracies have bootstrap $95\%$ CIs (omitted for compactness; \texttt{current} is $[77.0,\,82.0]$ / $[75.3,\,79.3]$ / $[74.2,\,79.5]$ across the three benchmarks); pipeline decisions are deterministic so no seed variance applies.}
\end{table}

\textbf{Reading Table~\ref{tab:prod-evolution}.}
\emph{(i)} Same rule content in Python ($79.6\%$ / $0.50$ Rej-F1) beats the same content as an Opus 4.6 prompt ($69.8 \pm 0.1\%$ / $0.17$): a $9.8$pp accuracy gap and $\sim 3{\times}$ Rej-F1 gap in favor of compilation.
\emph{(ii)} The pipeline outperforms all LLM baselines: minimal-prompt Opus ($43.8\%$), and the historical pure-LLM ($72.7\%$) by $\sim 7$pp.
\emph{(iii)} Drift-response rules act as regularizers, improving held-out and drift splits at a small training-benchmark cost (peak $81.1 \to$ current $79.6$).

\subsection{LLM Fallback Ablation: Challenging the Sweet Spot Hypothesis on Skewed-Base-Rate Tasks}
\label{sec:prod-c}

A common design pattern pairs a rule cascade with an LLM safety net: unresolved cases route to an LLM, which is expected to recover the long tail that rules cannot deterministically classify. Our pipeline contains exactly this gate (\texttt{pipeline.py}, lines 857ff) but in practice it is never invoked, because an earlier rule defaults all unresolved cases to ``approve'' with a fixed reason string (``no reliable voice evidence + AuditPlatform intact, default to allocation''). We test whether activating the fallback would help.

\textbf{Design.} The \texttt{C} variant replaces the default-pass assignment at lines 843--855 with an \texttt{if decision is None: pass}, allowing cases to fall through to the existing LLM-fallback block. The high-reject-code path and weak-claim path remain unchanged. This is the minimal change that turns the dormant safety net on.

\textbf{Results.} Table~\ref{tab:prod-c} reports the three-benchmark comparison, the realized LLM call rate (counted from audit-output reasons), and the shift in false rejections.

\begin{table}[t]
\centering
\caption{C ablation: activating the LLM fallback at $13$--$20\%$ call rate degrades accuracy on all three benchmarks, and degrades \emph{more} under distribution shift. All LLM calls are to Claude Opus 4.6.}
\label{tab:prod-c}
\small
\begin{tabular}{lccccc}
\toprule
Benchmark & \texttt{current} & \texttt{C} & $\Delta$ accuracy & LLM call rate & $\Delta$ false rejections \\
\midrule
\texttt{benchmark\_1000} (training) & $79.6\%$ & $78.6\%$ & $-1.0$pp & $13.3\%$ & $44 \to 59$ \\
\texttt{new\_2000} (held-out)        & $77.3\%$ & $74.4\%$ & $-2.9$pp & $15.6\%$ & $59 \to 140$ \\
\texttt{recent\_week} (drift)        & $76.8\%$ & $72.5\%$ & $-4.3$pp & $19.7\%$ & $86 \to 96$ \\
\bottomrule
\end{tabular}
\end{table}

\begin{figure}[t]
\centering
\includegraphics[width=0.82\linewidth]{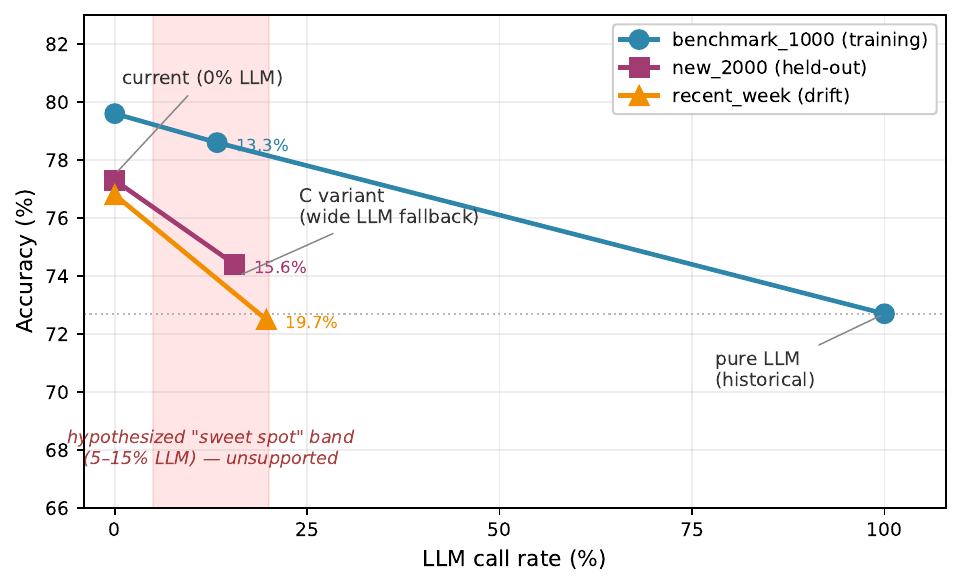}
\caption{LLM call rate vs.\ accuracy on three benchmarks. All three lines decrease monotonically from \texttt{current} ($0\%$ LLM) to the \texttt{C} variant ($13$--$20\%$ LLM), and the decrease is steeper under distribution shift (drift slope $-0.22$pp per percentage-point of LLM rate) than on training ($-0.08$pp). The historical pure-LLM baseline ($72.7\%$ on \texttt{benchmark\_1000}) is shown for reference; the hypothesized $5$--$15\%$ ``sweet spot'' band is unsupported by the observed monotone trend.}
\label{fig:c-monotone}
\end{figure}

Figure~\ref{fig:c-monotone} plots the table. The accuracy loss is driven entirely by false rejections: the LLM systematically prefers ``reject'' over ``approve,'' violating the $\sim$75\% pass base rate, while the rule cascade's default-pass branch is a calibrated prior that 22 days of EISR have tuned. The gap widens under drift because more cases fall through to the uncalibrated LLM. On this skewed-base-rate task, compiled rules outperform LLM fallback at every mixing ratio tested; we do not claim this generalizes to all decision tasks.

\subsection{Scope of the Compilation Claim and Cross-Regime Behavior}
\label{sec:compilation-bridge}

\emph{Within the production regime}, switching from skill-level prompt to compiled Python (with production-driven extensions accumulated over the 22-day deployment) yields the $9.8$\,pp gap reported in \S\ref{sec:prod-evolution}; \texttt{pipeline.py} is an independent Python implementation with production-driven extensions, so the gap bundles form with engineering, and isolating the form-only component is a measurement gap we flag as future work. \emph{Out of regime}, the pipeline reaches only $74.1\%$ on the $139$-case skill-level held-out, below v8 ($77.2\%$), Auto-EISR ($80.6\%$), and B1b Opus ($82.7\%$): the calibrated prior is engineered against the production traffic distribution and does not transfer cross-regime, so compilation is a regime-specific deployment choice rather than a universal dominator (Appendix~\ref{app:forensic}).

\section{Experiments}
\label{sec:exp}

\emph{Scope.} This section evaluates skill-level rules (Markdown prompt executed by an LLM); the pipeline-level (compiled Python) is in Section~\ref{sec:production}.

\subsection{Setup}

Logistics damage liability audit at a major Chinese logistics carrier: given a waybill's AuditPlatform audit data, the agent decides approve/reject and assigns a category.\footnote{GLM-5 has higher skip rate ($\sim$25\%) due to API instability. Accuracy is computed over valid responses only.}
555 real audit trajectories from two expert auditors yield 476 structured records (85.8\%); train/test split 444/111 (seed=42); EISR-refined policy: 62 rules, 4 paths, 41 situations.
Six LLMs evaluated: GLM-5, Kimi-K2.5, Qwen3.5-plus, MiniMax-M2.5, Claude Opus 4.6, Haiku 4.5.
Four conditions: \textbf{B1b} (zero-shot), \textbf{B5} (5-shot ICL), \textbf{B7} (v1 Skills, no EISR), and \textbf{v8} (EISR-refined).
We report on the 111-waybill validation pool (used during EISR; baselines never exposed) and two held-out sets (40-case and 139-case, disjoint from the pool; Appendix~\ref{app:held-out}).

\subsection{Main Results}
\label{sec:main-results}

\begin{table}[t]
\centering
\caption{\textbf{In-sample results.} Action / category accuracy (\%) on the 111-waybill validation pool across 5 models and 4 conditions. v8 rules were refined against this pool; other rows were not. In-sample numbers do not constitute out-of-distribution generalization evidence; see Table~\ref{tab:auto-eisr} for held-out results. Shaded rows form the ``surface knowledge ceiling'' ($\sim$70\%). Opus 4.6 is a frontier model (Claude); others are smaller Chinese LLMs.}
\label{tab:baseline}
\small
\begin{tabular}{lcccccc}
\toprule
\textbf{Condition} & \textbf{GLM-5} & \textbf{Kimi} & \textbf{Qwen} & \textbf{MMx} & \textbf{Opus 4.6} & \textbf{Avg} \\
\midrule
\rowcolor{gray!10} B1b: No rules & 72.6/40.6 & 56.7/31.7 & 75.2/36.7 & 73.9/35.1 & 73.3/43.6 & 70.3/37.5 \\
\rowcolor{gray!10} B5: Few-shot  & 75.6/56.7 & 66.7/36.2 & 80.0/55.2 & 66.0/34.0 & 78.3/51.9 & 73.3/46.8 \\
\rowcolor{gray!10} B7: v1 Skills & 66.7/44.4 & 72.5/32.1 & 70.9/44.5 & 70.8/32.1 & 64.9/44.1 & 69.2/39.4 \\
\textbf{v8: EISR} & \textbf{84.5/69.0} & \textbf{78.4/49.5} & \textbf{80.2/54.1} & \textbf{71.6/46.8} & \textbf{79.6/55.6} & \textbf{78.9/55.0} \\
\bottomrule
\end{tabular}
\end{table}

Table~\ref{tab:baseline} reveals a \emph{layered knowledge effect}: three non-refinement approaches cluster around $\sim$70\% (B1b 70.3\%, B5 73.3\%, B7 69.2\%), while EISR-refined v8 lifts performance to 78.9\% on average, outperforming each model's best non-refinement baseline. Category accuracy shows an even larger gap ($55.0\%$ vs $46.8\%$ best non-refinement). The subsections below summarize six follow-up analyses; detailed tables and figures for each are in the corresponding appendices.

\textbf{Authority displacement.}
On the 139-case held-out, handing strong executors (Opus, Haiku) unrefined v1 rules \emph{reduces} accuracy by $7$--$9$pp relative to zero-shot, while weak executors (Kimi) gain $+22.8$pp. GLM-5 is anomalous (helped despite high B1b). We attribute this to models deferring to explicit rules even when their own reasoning is superior; EISR-refined rules close most of this gap (Table~\ref{tab:displacement} in Appendix~\ref{app:authority}).

\textbf{Training convergence.}
Both Human-EISR and Auto-EISR show non-monotonic per-round accuracy; the R5 crash ($90\% \to 70\%$) surfaced the encoding-state trap (Trap 1) and was the most informative training event. Full trajectory in Appendix~\ref{app:convergence}.

\textbf{Input representation.}
Compact structured API input (810 chars) consistently outperforms raw DOM scrape (6K chars) by $+8.4\%$ average action accuracy across 5 models, isolating the value of Phase 2's platform layer (Appendix~\ref{app:input-repr}).

\textbf{External validation.}
A previously unused Opus instance reached $85.6\%$ action / $67.6\%$ category on the 111-pool with v8 rules, confirming cross-model transfer.

\textbf{Error analysis.}
v8's remaining errors: ambiguous ground truth ($\sim$40\%), rare path combinations ($\sim$35\%), implicit visual evidence ($\sim$25\%).

\textbf{Phase 4 validation (v11).}
A practitioner-redesigned policy (``v11'') that bypassed EISR---replacing 62 rules with a 4-rule decision tree---regressed to $66.8\%$ (validation) and $67.5\%$ (held-out), below every baseline including B1b. This confirms that every policy change needs a regression gate; the difference between Auto-EISR's success and v11's failure is precisely that Auto-EISR enforces this gate mechanically (Appendix~\ref{app:v11}).

\subsection{Automating EISR}
\label{sec:auto-eisr}

We implement \textbf{Auto-EISR}, an LLM-driven Phase 3 variant: two prompted agents (Executor: Opus 4.6; Diagnose/Refine: Kimi-K2.5) with a pre-flight validator and regression gate ($>$2\% drop $\to$ rollback). Three 8-round runs (seeds 42/43/44) each produce a best-snapshot rule set (Appendix~\ref{app:auto-eisr}).

\begin{table}[t]
\centering
\caption{Held-out accuracy (\%), clean of EISR exposure. Top: 5-model avg on 40-case held-out. Bottom: single-executor on 139-case extended held-out. Auto-EISR: mean $\pm$ std across 3 seeds; deployed configuration disables regression gate (Appendix~\ref{app:safeguard-ablation}).}
\label{tab:auto-eisr}
\small
\begin{tabular}{llcccc}
\toprule
\textbf{Executor regime} & \textbf{Metric} & \textbf{B1b} & \textbf{v1} & \textbf{Human-v8} & \textbf{Auto-EISR (3-seed)} \\
\midrule
\multirow{2}{*}{Strong 5-model avg (40 held-out)} & Action   & --   & 76.4 & 75.1 & 76.4 $\pm$ 3.6 \\
                                      & Category & --   & 36.4 & \textbf{52.2} & 38.8 $\pm$ 3.8 \\
\midrule
\multirow{2}{*}{Opus 4.6 (139 held-out)}     & Action   & \textbf{82.7} & 74.1 & 77.2 & 80.6 $\pm$ 4.5 \\
                                              & Category & 36.7 & 36.7 & 48.0 & \textbf{50.4} \\
\midrule
\multirow{2}{*}{Haiku 4.5 (139 held-out)}     & Action   & \textbf{84.2} & 77.0 & 75.0 & 81.3 $\pm$ 1.9 \\
                                              & Category & 37.4 & 35.3 & 50.0 & \textbf{50.1} \\
\bottomrule
\end{tabular}
\end{table}

Auto-EISR matches Human-EISR on action accuracy: on the 139-case held-out, Opus Auto-EISR $80.6 \pm 4.5\%$ vs v1 $74.1\%$ and Human-v8 $77.2\%$ (McNemar $p{=}0.029$, uncorrected; does not survive Bonferroni $\alpha'{=}0.025$); Haiku $81.3 \pm 1.9\%$ vs v1 $77.0\%$ ($p{=}0.071$, n.s.).
Category accuracy is the most reliable contribution: $+13$--$14$pp over B1b ($50\%$ vs $37\%$).
The deployed configuration disables the regression gate ($+5$pp on category; Appendix~\ref{app:safeguard-ablation}).

\textbf{Cost.}
One 8-round run: 2--3 hours, \$5--\$10 API cost vs $\sim$70 expert-hours for Human-EISR, re-runnable at zero expert time as distributions drift (Appendix~\ref{app:per-seed-stats}).

\section{Cross-Domain Transferability: A Public-Benchmark Probe}
\label{sec:cross-domain}

The damage-audit study above is single-domain. We probe cross-domain transfer at two levels: a detailed Human-EISR study on LegalBench's hearsay task~\citep{guha2023legalbench} (94 binary fact patterns under FRE~801; 30-case iteration / 64-case held-out), and Auto-EISR on four public benchmarks (three LegalBench tasks + BPIC~2012, Appendix~\ref{app:cross-domain-extended}). Phase~0--1 is bypassed; Phase~3's refinement loop applies without modification. \textbf{Setup.} Three executors (Opus 4.6 / Haiku 4.5 / Kimi K2.5) $\times$ three conditions \{\emph{B1b} zero-shot; \emph{v1} hand-authored FRE~801; \emph{v\_EISR} two refinement rounds\}, $T{=}0$.

\begin{table}[h]
\centering
\small
\caption{LegalBench hearsay (64-case held-out): accuracy and Phase 3 lift. Bold marks the best per row. McNemar exact $p$ for v\_EISR vs.\ v1.}
\label{tab:legalbench-hearsay}
\begin{tabular}{lcccc|c}
\toprule
Executor & B1b & v1 & v\_EISR (R1) & \textbf{v\_EISR (R2)} & McNemar $p$ \\
\midrule
Opus 4.6           & $82.8\%$ & $89.1\%$ & $85.9\%$ & $\mathbf{92.2\%}$ & $0.69$ \\
Haiku 4.5          & $79.7\%$ & $85.9\%$ & $87.5\%$ & $\mathbf{93.8\%}$ & $0.13$ \\
Kimi K2.5          & $79.7\%$ & $76.6\%$ & $81.2\%$ & $\mathbf{90.6\%}$ & $\mathbf{0.049}$ \\
\midrule
mean               & $80.7\%$ & $83.9\%$ & $84.9\%$ & $\mathbf{92.2\%}$ & --- \\
\bottomrule
\end{tabular}
\end{table}

\textbf{Observations.} \emph{(i)} v\_EISR (R2) lifts every executor above $90\%$ ($+8$--$13$pp; Kimi $p{=}0.049$). \emph{(ii)} Round~1 overshoots (``Not introduced to prove truth'' $-36$--$50$pp); Round~2 recovers to $86$--$100\%$, mirroring Auto-EISR's gating trajectory; v\_EISR (R2) helps every executor regardless of v1's starting point (Appendix~\ref{app:legalbench}). \emph{(iii)} The separate Auto-EISR evaluation (Appendix~\ref{app:cross-domain-extended}) shows positive mean $\Delta_E$ on 3/4 tasks across three executors~\citep{vandongen2012bpic}. $N{=}64$ limits per-cell power; rigorous cross-domain evaluation remains necessary.

\section{Conclusion}

We presented Trace2Policy, a framework that extracts and iteratively refines expert decision rules from behavior traces. EISR surfaces deep knowledge invisible to one-shot extraction; Auto-EISR reproduces this at ${\sim}1/100$ the cost. Across a 22-day deployment (3{,}349 cases) and four public benchmarks, compiled EISR-refined rules consistently outperform both unrefined rules and direct LLM prompting. Within a compliance-sensitive vertical and within the deployment regime, rule quality yields larger accuracy returns than upgrading models---an observation, not a universal prescription.

\textbf{Limitations.} Our evaluation covers two domains (damage-audit primary, LegalBench/BPIC probe-level) with modest held-out sizes (40--297 cases); systematic ablation of individual EISR components (cluster count $K$, gate sensitivity) and an isomorphic-compilation experiment isolating form from engineering remain open. Auditor labels in deployment are observed under agent recommendation, so anchoring may inflate headline accuracies; extended limitations are in Appendices~\ref{app:discussion} and~\ref{app:limitations}.

\textbf{Future work.} Three directions stand out: (1)~component-level ablation to isolate EISR's active ingredients, (2)~broader multi-domain validation with domain-specific baselines and larger sample sizes, and (3)~a blind sub-study (auditor labels collected without exposure to agent output) and isomorphic compilation experiments to cleanly bound the form-vs-engineering gap.

\begin{ack}
[Redacted for double-blind review.]
\end{ack}

\bibliography{references}
\bibliographystyle{plainnat}

\newpage
\appendix

\section{Extended Discussion}
\label{app:discussion}

\textbf{The two studies, read together: compilation is the mechanism.}
Section~\ref{sec:production} (pipeline-level) and Section~\ref{sec:exp} (skill-level) together produce a more nuanced picture than either in isolation. Within the production regime, the form of the rules matters substantially: on \texttt{benchmark\_1000} the same content reaches $71.0\%$ as a prompt and $79.6\%$ as Python, and widening LLM fallback monotonically degrades accuracy on all three production benchmarks. Cross-regime evidence (\S\ref{sec:compilation-bridge}) shows the pipeline's advantage does not automatically transfer: on the $139$-case skill-level held-out the pipeline reaches only $74.1\%$, below B1b, Auto-EISR, and v8. Compilation is a first-class design axis but complementary to semantic refinement. Auto-EISR automates the semantic-design half; compilation is a human-in-the-loop step that deserves its own tooling research.

\textbf{Why every phase still matters.}
(1) Platform layer (Phase 2): $\sim$8\% accuracy drop without it (Appendix~\ref{app:input-repr}).
(2) Surface-knowledge ceiling at $\sim$70\% is pool-specific; held-out v1 yields $\sim$74--77\%.
(3) Authority displacement: v1 rules cost $7$--$9$pp on strong executors, add $+22.8$pp on weak (Kimi). Compilation bypasses displacement entirely.
(4) v11 regression: compilation without EISR discipline regresses; with discipline it wins.

\textbf{EISR as the framework's keystone.}
EISR refines initial policies (Phase 2$\to$3) \emph{and} validates production changes (Phase 4). The v11 regression shows even experienced practitioners cannot reliably predict which changes preserve accuracy. Structured error diagnosis enables discovery of deep knowledge (trap rules) that one-shot analysis cannot extract.

\textbf{When is Trace2Policy applicable?}
Tasks where: (1) experts make repeated decisions using software systems, (2) decisions follow implicit but systematic rules, and (3) ground truth is available. Insurance claims, compliance audits, and quality inspections fit this profile.

\section{Limitations and Broader Impacts}
\label{app:limitations}

(1) Two-domain validation: damage-audit (primary) plus LegalBench hearsay ($N{=}64$). Broader multi-domain sweeps would strengthen generalizability.
(2) Held-out sizes (40 and 139 cases) leave some McNemar comparisons underpowered; larger draws would tighten intervals.
(3) Two executor regimes (Opus-4.6 + Haiku-4.5); the capability continuum is not systematically swept.
(4) The 555-record dataset is relatively small.
(5) Text-only evidence; extending to visual evidence (damage photos) is important.
(6) Online validation is preliminary (5 waybills).
(7) Auto-EISR is trained against action labels only; category supervision is a natural next step.
(8) The safeguard ablation (Table~\ref{tab:ablation-safeguards}) covers a limited configuration space.
(9) \emph{Ablation coverage.} The current evaluation does not include ablation studies isolating individual EISR components (e.g., number of trace clusters, gating mechanism, rule extraction threshold). This limits the strength of our methodological contribution claims and we plan to address this gap in future work.
(10) \emph{Refinement signal limitations.} EISR's rule evolution is constrained by the 111-case validation set's distributional coverage, which may not guarantee out-of-distribution robustness. Future work: online refinement, bootstrap stability detection, and rule complexity penalties.
(11) \emph{Auditor anchoring on the natural ground truth.} We characterize human auditor decisions as ``ground truth at zero marginal cost'' (\S\ref{sec:flywheel}). This framing assumes auditor labels are independent of the agent's reference answers, but in deployment auditors see the agent's recommendation before committing their own decision. Automation-bias and decision-anchoring effects~\citep{mosier1996automation,goddard2012automation,bond2018automationbias} could systematically bias the labels toward agent agreement, which would inflate the headline pipeline accuracies (Table~\ref{tab:prod-evolution}) by an unknown amount. The benchmarks reported in this paper were collected under the production workflow and inherit this risk; a blind sub-study (auditor labels collected without exposure to agent output) is the right way to bound the effect. We flag this as the most important methodological caveat for any future deployment that uses the natural-flywheel signal as evaluation rather than just as a refinement trigger.

\section{Theoretical Analysis}
\label{sec:theory}

\textbf{Formal setup.}
Let $\mathcal{X}$ be the input space, $\mathcal{Y} = \mathcal{A} \times \mathcal{C}$ the label space (action $\times$ category).
A policy $S = \{(c_i, y_i, p_i)\}$ is a prioritized rule list.
Given validation set $V$ of size $n$, the error set is $E_S = \{k : S(x_k) \neq y_k\}$.

\begin{figure}[h]
\centering
\begin{tikzpicture}
\begin{axis}[
    width=\columnwidth,
    height=5.2cm,
    xlabel={Round},
    ylabel={In-sample batch action acc.\ (\%)},
    xmin=0.5, xmax=8.5,
    ymin=40, ymax=105,
    xtick={1,2,3,4,5,6,7,8},
    xticklabels={R1,R2,R3,R4,R5,R6,R7,R8},
    ytick={50,60,70,80,90,100},
    legend style={at={(1.02,0.5)}, anchor=west, font=\scriptsize, draw=gray!50},
    grid=major, grid style={gray!20},
    tick label style={font=\small}, label style={font=\small},
]
\addplot[gray!60!black, thick, mark=square*, mark size=2pt, mark options={fill=gray!60!black}] coordinates {
    (1,90) (2,85) (3,80) (4,90) (5,70) (6,90) (7,80) (8,100)
};
\addlegendentry{Human-EISR}
\addplot[blue!70!black, thick, mark=*, mark size=2pt, mark options={fill=blue!70!black}] coordinates {
    (1,80) (2,60) (3,60) (4,93.3) (5,66.7) (6,73.3) (7,80) (8,66.7)
};
\addlegendentry{Auto-EISR seed 42}
\addplot[red!70!black, thick, mark=triangle*, mark size=2pt, mark options={fill=red!70!black}] coordinates {
    (1,66.7) (2,80) (3,73.3) (4,93.3) (5,66.7) (6,66.7) (7,80) (8,93.3)
};
\addlegendentry{Auto-EISR seed 43}
\addplot[green!60!black, thick, mark=diamond*, mark size=2pt, mark options={fill=green!60!black}] coordinates {
    (1,66.7) (2,93.3) (3,66.7) (4,80) (5,86.7) (6,93.3) (7,66.7) (8,86.7)
};
\addlegendentry{Auto-EISR seed 44}
\node[font=\tiny, text=green!40!black] at (axis cs:3,60) {$\times$};
\node[font=\tiny, text=green!40!black] at (axis cs:6,87) {$\times$};
\addplot[gray, thin, dashed, domain=0.5:8.5] {75.1};
\node[font=\tiny, text=gray!60!black, anchor=west] at (axis cs:0.7,77) {Human-EISR held-out avg ($75.1\%$)};
\end{axis}
\end{tikzpicture}
\caption{In-sample per-round action accuracy for Human-EISR (one run) and Auto-EISR (three seeds). $\times$ marks regression-gate rollbacks.}
\label{fig:convergence}
\end{figure}
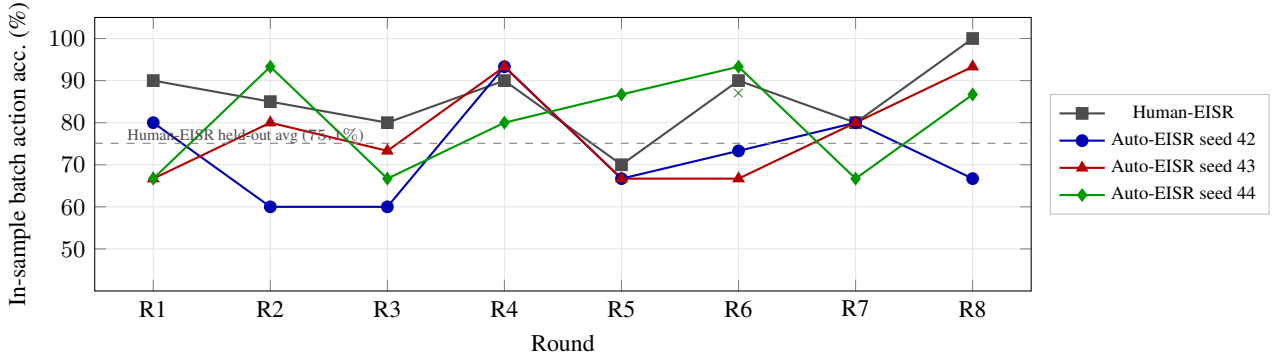

\textbf{EISR as version space narrowing.}
Following Mitchell~\citep{mitchell1982generalization}, define the version space $\mathcal{V}_t$ as all policies consistent with correctly-classified cases after round $t$.
Each EISR round uses errors as counterexamples to shrink $\mathcal{V}_t$.
When refinement introduces no regressions, $\mathcal{V}_{t+1} \subseteq \mathcal{V}_t$; in practice, regressions can temporarily expand it, explaining the non-monotonic convergence in Figure~\ref{fig:convergence}.

\textbf{Why structured diagnosis helps.}
Let $m$ be the number of distinct root causes and $|E_S|$ the total errors ($m \ll |E_S|$).
If each fix attempt has regression probability $\rho$, expected regressions are $\rho \cdot m$ (structured) vs.\ $\rho \cdot |E_S|$ (unstructured).
Empirically, R5 had $|E_S|=6$ errors from $m=2$ root causes; two targeted fixes in R6 resolved 4 of 6 errors.

\textbf{Sample complexity.}
To detect errors with rate $\varepsilon$ at confidence $1-\delta$, we need $n \geq \ln(1/\delta)/\varepsilon$ validation cases.
For $\varepsilon=0.10, \delta=0.05$: $n \geq 30$.
Our validation sets (10--20 per round) suffice for high error rates but may miss rare errors.

\textbf{Generalization and held-out behaviour.}
Under PAC assumptions with $|\mathcal{S}_K| \approx 100$ and $n=111$, the naive bound is $\sim\sqrt{\ln|\mathcal{S}_K|/n} \approx 20\%$, consistent with the empirical gap between terminal in-sample accuracy ($\sim$95\%) and held-out ($\sim$75\%).

\section{Positioning Summary}
\label{app:positioning}

\begin{table}[h]
\centering
\caption{Positioning of Trace2Policy among related approaches. ``Training-time gate'' means each proposed refinement is validated against a held-out batch before being committed during refinement (independent of inference-time configuration). DSPy/MIPRO use compiler-style validation; AI Scientist v2, FunSearch, and Auto-EISR (ours) additionally enforce per-step regression checks. Note: in our deployed Auto-EISR configuration, the inference-time gate is disabled to optimize category accuracy (App.~\ref{app:safeguard-ablation}); the training-time gate is always on and is the structural mechanism reported here.}
\label{tab:positioning}
\small
\begin{tabular}{lcccc}
\toprule
\textbf{Approach} & \textbf{Learns} & \textbf{Refines via} & \textbf{Interpretable} & \textbf{Training-time gate} \\
\midrule
GUI Agents & Operations (how to click) & -- & No & -- \\
Voyager / SAGE & Code skills (programs) & Iter.\ LLM + env feedback & Partially & No \\
Self-Refine / Reflexion & Outputs / memory & Self-feedback & No & No \\
TextGrad / OPRO & Prompts & Text gradients & No & No \\
AgentRefine & Behavior & Env.\ feedback & No & No \\
DSPy / MIPRO & Prompt programs & Compiler-style search & Partially & Validation set \\
AI Scientist v2 & Experiment variants & Tree search + reviewer & Partially & Yes \\
FunSearch & Code (programs) & Evolutionary + evaluator & Yes & Yes \\
Process Mining & Workflows (what) & -- & Yes & -- \\
\textbf{Trace2Policy} (Human-EISR) & \textbf{Decision rules (why)} & \textbf{Human error analysis} & \textbf{Yes} & \textbf{Manual} \\
\textbf{Trace2Policy} (Auto-EISR) & \textbf{Decision rules (why)} & \textbf{LLM Diag+Refine} & \textbf{Yes} & \textbf{Yes (training-time)} \\
\bottomrule
\end{tabular}
\end{table}

\section{Phase 0--1 Implementation Details}
\label{app:phases01}

\textbf{Agent Observer sensing layers.}
The system captures four event types: (1) window focus changes with timestamps, (2) mouse clicks with 400$\times$300 pixel region screenshots centered on the click point, (3) keyboard and text input including paste events, and (4) browser events (network requests, page navigations, DOM mutations).
Raw event streams are segmented into task-level trajectories using \emph{business anchors}---domain identifiers extracted from clipboard or input events (e.g., waybill numbers matching the carrier-specific format).
Privacy is handled through password masking, configurable domain blacklists, local-only storage, and explicit opt-in.

\textbf{VLM structurization output.}
For each trajectory, the VLM produces: systems consulted (platforms and order), key observations (inspected data fields), reasoning chain (logical steps), decision and justification, and supporting evidence.
The 85.8\% success rate (476/555) reflects failures from incomplete recordings or system errors during the original session.

\section{Phase 4: Production Data Flywheel Details}
\label{app:flywheel}

The production deployment creates a natural feedback loop.
The agent provides reference answers; human auditors review every case as part of their unchanged workflow, generating ground truth at zero marginal annotation cost.
Errors are automatically collected and clustered; when similar errors accumulate beyond a threshold ($\geq 5$ cases with the same root cause), an incremental EISR round triggers.

\textbf{Cold-start vs.\ incremental EISR.}
Cold-start (Phase 2--3) uses historical records, runs 5--8 EISR rounds with major rule additions, and validates on a hold-out test set.
Incremental (Phase 4) uses daily production errors, typically requires a single iteration modifying 1--2 rules, and validates via regression on historical cases.

\textbf{Progressive trust.}
As accuracy improves, high-confidence cases (agent-human agreement) can gradually bypass human review, releasing human capacity for novel or ambiguous cases.

\subsection*{Production Pipeline (\texttt{pipeline.py}) Structure}
\label{app:pipeline-structure}

The deployed pipeline is $1{,}193$ lines of Python (including imports, logging, and I/O), of which roughly $600$ lines constitute the decision rule cascade \texttt{process\_one}. Rules are applied in a strict priority order; once a rule assigns a decision, later rules are skipped (\texttt{if decision is None:} guards). The major layers, in order of invocation, are:

\begin{enumerate}[nosep,leftmargin=*]
    \item \textbf{Pre-pass codes} (lines ${\sim}370$--$420$): a small set of specific liability codes (anonymized; \emph{misreport / non-damage} category) that unambiguously indicate approval; these short-circuit the cascade.
    \item \textbf{Normative codes with audit evidence} (lines ${\sim}430$--$500$): an enumerated set of normative codes (anonymized), combined with system-detected packaging or documentation nonconformity, yield rejection.
    \item \textbf{AuditPlatform arbitration} (lines ${\sim}499$--$600$): the single largest block, handling the common AuditPlatform-packaging-state vs.\ RouteSystem-voice-evidence conflict. Contains the headline rule revision documented in the paper: ``AuditPlatform=damaged $+$ receiver-voice=intact $\to$ trust receiver voice,'' which substantially reduced false rejections in the affected cluster.
    \item \textbf{Voice-evidence resolution} (lines ${\sim}600$--$700$): when AuditPlatform is intact but RouteSystem contains phone-call statements about outer-packaging state, classify by which party's voice is recorded.
    \item \textbf{Drift-response rules} (lines ${\sim}700$--$843$, added 2026-04-16 through 2026-04-22): five rules (internally N1, N2, N3, and two solution patches) that cover patterns introduced by the \texttt{recent\_week} benchmark: appeal-class signal mismatches, DetectorService detector thresholding, and RouteSystem-call approximations.
    \item \textbf{Default-pass branch} (lines $843$--$855$): cases with a ``some evidence but nothing dispositive'' profile---AuditPlatform intact and no high-reject-code signal---default to approval with the reason string ``no reliable voice evidence $+$ AuditPlatform intact, default to allocation.'' This branch handles the majority of cases; it is the branch the C variant replaces.
    \item \textbf{LLM fallback gate} (lines $857$ onward): \emph{dormant in the deployed pipeline}. Triggers only if \texttt{decision is None} after all above branches; in practice, step 6's default-pass assignment always fires first. The C experiment re-routes step 6's cases through this gate.
    \item \textbf{Path-and-category assignment} (lines ${\sim}950$--$1040$): given a decision and code, assign E/C/A/BD decision path and one of $17$ category codes. This layer is independent of the approve/reject decision and applies even to LLM-routed cases.
\end{enumerate}

Rule edits across the 22 days are tagged in the source with the edit date and, where applicable, a failure-cluster rationale. Representative comments (translated): ``rule contributed roughly two-thirds of false rejections in the affected cluster''; ``kept as hard-evidence exception: RouteSystem ImageAnalysis collection explicitly marks outer packaging damaged''; ``strategy: prefer over-rejection (can be manually re-adjudicated) over under-rejection (approval is irreversible).''

\subsection*{LLM Call Rate: Why 0\%, Not ``<1\%''}
\label{app:llm-rate-zero}

Early versions of the pipeline's skill documentation described LLM usage as ``<1\%'' conditional fallback. Our direct measurement of the current deployed state finds $0\%$: the LLM gate (layer 7 above) is never reached because the default-pass branch (layer 6) assigns a decision to every case that survives layers 1--5. We verified this by parsing the reason field of every waybill in all three benchmark runs and matching it against the closed set of rule-generated reason templates; no run produces a free-form LLM-authored reason string under \texttt{current} pipeline conditions. The ``<1\%'' claim reflected an earlier pipeline version before the default-pass branch was promoted above the LLM gate; it is now stale. The $0\%$ measurement is what the C experiment's widening targets.

\subsection*{C Experiment: Measurement and the Proxy Trap}
\label{app:c-measurement}

Computing the LLM call rate in the C variant required care because the benchmark script does not log LLM invocations. We reconstruct it from per-waybill reason strings. A waybill counts toward LLM call rate if its reason is either (a) a free-form string that does not match any of the 33 known rule-template prefixes (LLM-success), (b) a specific sentinel string (``AuditPlatform intact $+$ no hard evidence, LLM rejection overridden'') indicating an LLM-proposed rejection was post-hoc overridden by the AuditPlatform-intact safety rule (LLM-covered), or (c) a different sentinel (``default-pass'') emitted when the LLM call raised an exception or returned non-JSON (LLM-fail). The totals reported in Table~\ref{tab:prod-c} are the sum of these three categories.

The first C run produced an anomalous $13.5\%$ ``routed but zero successful'' rate: every LLM call returned HTTP 502. The cause was the corporate HTTP proxy, which Claude Code's system proxy settings had configured for all outbound traffic. The internal model hub is inside the same network and does not accept proxied traffic; direct requests succeed. Rerunning the C experiment with \texttt{http\_proxy=} and \texttt{https\_proxy=} explicitly unset produced the results reported in the main text. We note this because the initial ``C $\approx$ current on benchmark\_1000'' observation (first run) was an artifact of broken LLM calls falling back to approval; once the proxy was fixed, the monotone-decreasing pattern became clear. This is a generic production-pipeline-evaluation pitfall and worth flagging for practitioners.

\subsection*{Per-Benchmark Construction}
\label{app:benchmark-construction}

\texttt{benchmark\_1000} (987 waybills, 2026-03-30): the training benchmark, drawn from audited waybills during Phase 3 cold-start EISR and stable across the 22-day window. Ground truth is the final human auditor decision, collected via the same natural workflow the deployed pipeline augments.

\texttt{new\_2000} (1{,}462 waybills, 2026-03-31): a held-out benchmark drawn from audited waybills immediately before the 22-day evolution period began; no rule edit in the window could have specifically targeted these cases. Used as the primary held-out generalization measurement.

\texttt{recent\_week} (894 waybills, 2026-04-16): drawn from the most recent week of production traffic at the time the drift-response rules were authored. By construction, rule edits 4-16 through 4-22 were informed by patterns in this set, so \texttt{recent\_week} is best read as a drift-\emph{exposure} benchmark rather than a clean held-out; its improvement between \texttt{bak\_v18} ($75.1$) and \texttt{current} ($76.8$) is partially in-sample. We interpret conservatively: the held-out number to lean on is \texttt{new\_2000} ($77.0 \to 77.3$, $+0.3$pp), and the drift number provides a direction check rather than an unbiased generalization estimate.

\section{Forensic Analysis: Cross-Regime Gap on the 139 Held-out}
\label{app:forensic}

Case-level data supporting the analysis in \S\ref{sec:compilation-bridge}. All numbers are from the pipeline run tagged 2026-04-23 14:53:45 on the $139$-case skill-level held-out.

\subsection*{Error Decomposition}

\begin{table}[h]
\centering
\small
\begin{tabular}{lrr}
\toprule
 & False Neg ($n{=}23$) & False Pos ($n{=}13$) \\
\midrule
Recovered by v8 or Auto-majority & $22$ & $8$ \\
Shared error (v8 \& Auto also fail) & $1$ & $5$ \\
\midrule
Compilation-gap share & $\mathbf{96\%}$ & $62\%$ \\
\bottomrule
\end{tabular}
\caption{False negatives are almost entirely recoverable by skill-level methods ($22/23$): the rule content exists but is not compiled into \texttt{pipeline.py}. False positives are split between pipeline-specific over-rejection ($8$) and ambiguous edges also missed by skill methods ($5$).}
\label{tab:forensic-decomp}
\end{table}

\subsection*{Path and Sub-Branch Distribution}

\begin{table}[h]
\centering
\small
\begin{tabular}{lrrrr}
\toprule
 & \multicolumn{2}{c}{\texttt{benchmark\_1000}} & \multicolumn{2}{c}{$139$ held-out} \\
\cmidrule(lr){2-3}\cmidrule(lr){4-5}
Path & Share & Acc & Share & Acc \\
\midrule
A (packaging dispute)       & $86.2\%$ & $80.3\%$ & $74.1\%$ & $74.8\%$ \\
C (insufficient cushioning) & $1.6\%$  & $75.0\%$ & $\mathbf{10.8\%}$ & $66.7\%$ \\
E (timeliness)              & $9.8\%$  & $78.4\%$ & $13.7\%$ & $78.9\%$ \\
BD (claim)                  & $2.0\%$  & $60.0\%$ & $1.4\%$  & $50.0\%$ \\
\midrule
Path-A ``\textit{$C$: intact $\to$ apportion}'' sub-branch & \multicolumn{2}{c}{$81.1\%$ ($n{=}766$)} & \multicolumn{2}{c}{$78.8\%$ ($n{=}85$)} \\
\bottomrule
\end{tabular}
\caption{Path share and accuracy on production vs.\ $139$. Path C is $7\times$ over-represented on $139$ with $8$pp lower accuracy. The sub-branch responsible for $19/23$ of the $139$ false negatives (Path A's ``intact-packaging $\to$ apportion'') runs $81.1\%$ in-regime and $78.8\%$ out-of-regime.}
\label{tab:path-breakdown}
\end{table}

\subsection*{Rejection F1}

\begin{table}[h]
\centering
\small
\begin{tabular}{lrrrr}
\toprule
Method / Regime & Acc & Rej-P & Rej-R & Rej-F1 \\
\midrule
pipeline @ \texttt{benchmark\_1000} & $79.6\%$\,[$77.0, 82.0$] & $0.69$ & $0.39$ & $0.50$\,[$0.43, 0.55$] \\
pipeline @ \texttt{new\_2000}       & $77.3\%$\,[$75.3, 79.3$] & $0.53$ & $0.29$ & $0.38$\,[$0.32, 0.43$] \\
pipeline @ \texttt{recent\_week}    & $76.8\%$\,[$74.2, 79.5$] & $0.59$ & $0.34$ & $0.43$\,[$0.36, 0.49$] \\
pipeline @ $139$ held-out           & $74.1\%$\,[$66.9, 81.3$] & $0.41$ & $0.28$ & $\mathbf{0.33}$\,[$0.16, 0.49$] \\
\midrule
\multicolumn{5}{l}{\emph{Opus baselines @ \texttt{benchmark\_1000}, 3 seeds @ $T{=}0.3$}} \\
Opus zero-shot (B1b)                & $43.8 {\pm} 0.5\%$ & $0.23 {\pm} 0.01$ & $0.52 {\pm} 0.01$ & $0.32 {\pm} 0.01$ \\
Opus $+$ skills\_v8 prompt          & $69.8 {\pm} 0.1\%$ & $0.29 {\pm} 0.03$ & $0.12 {\pm} 0.02$ & $0.17 {\pm} 0.03$ \\
\midrule
\multicolumn{5}{l}{\emph{Skill-level @ $139$ held-out (Opus)}} \\
v1 (Opus)                           & $74.1\%$ & $0.46$ & $0.63$ & $0.53$ \\
v8 (Opus)                           & $77.2\%$ & $0.51$ & $0.90$ & $0.65$ \\
Auto-EISR 3-seed (Opus, deployed $-$gate)  & $80.6 {\pm} 4.5\%$ & $0.56 {\pm} 0.06$ & $0.79 {\pm} 0.06$ & $\mathbf{0.65 {\pm} 0.06}$ \\
Auto-EISR 3-seed (Haiku, deployed $-$gate) & $81.3 {\pm} 1.9\%$ & $0.57 {\pm} 0.03$ & $0.83 {\pm} 0.02$ & $\mathbf{0.67 {\pm} 0.02}$ \\
B1b (Opus, no rules)                & $82.7\%$ & $0.64$ & $0.56$ & $0.60$ \\
\midrule
always-\textit{approve} @ $139$     & $77.0\%$ & ---    & $0.00$ & $0.00$ \\
\bottomrule
\end{tabular}
\caption{Rejection precision, recall, and F1 with uncertainty. Pipeline CIs are bootstrap $95\%$ intervals (deterministic decisions; $n_{\text{boot}}{=}2{,}000$ over benchmark cases). Opus baselines and Auto-EISR report 3-seed mean $\pm$ std. The pipeline's rejection recall is universally $0.28$--$0.39$ across all four benchmarks; accuracy hides this because the rejection base rate is $23$--$26\%$ everywhere (trivial \textit{always-approve} scores $\sim\!77\%$). Crucially, the prompt-form of the \emph{same rules} (Opus $+$ skills\_v8) collapses on rejection F1 ($0.17 \pm 0.03$) relative to the compiled pipeline ($0.50$), even though its nominal accuracy ($69.8\%$) looks closer to the pipeline ($79.6\%$)---the apparent gap of $\sim\!10$pp in accuracy is a $\sim\!3\times$ gap in rejection F1.}
\label{tab:reject-f1}
\end{table}

\subsection*{The Named Missing Rule}

Skills\_v8 reasoning on the $22$ recovered false negatives consistently cites a pattern of the form: ``the appellant argues $X$, but the AuditPlatform audit record has explicitly determined $Y$ and the appellant has not provided new evidence that would overturn $Y$ $\to$ \textit{reject}.'' We refer to this as \emph{prior-audit precedence}. It is present as natural-language guidance in skills\_v8 but has not been compiled into \texttt{pipeline.py}'s Path~A ``intact-packaging $\to$ apportion'' branch. Its contribution on \texttt{benchmark\_1000} is below the $2$pp EISR discipline threshold; on the training-derived $139$ subset it would recover roughly $6$pp.

\subsection*{Confidence Calibration}

Pipeline confidence is not a reliable trigger for selective LLM fallback. On the $139$ cases: $72/103$ correct cases ($70\%$) receive ``high'' confidence, and $23/36$ incorrect cases ($64\%$) also receive ``high''; medium confidence splits similarly. A confidence-gated fallback is therefore not tractable on this signal; any hybrid routing would need an independent classifier, not pipeline self-assessment.

\section{LegalBench Hearsay Probe Details}
\label{app:legalbench}

This appendix documents the public-benchmark probe of \S\ref{sec:cross-domain}.

\textbf{Data and split.} LegalBench's \texttt{hearsay} task: $94$ test items with binary Yes/No labels and a categorical \texttt{slice} field over five hearsay categories. We took the full $94$ test items, deterministic-shuffled with seed $42$, and split into a $30$-case iteration set (used for EISR error analysis only) and a $64$-case held-out set. The held-out's slice distribution is: Standard hearsay $19$, Non-assertive conduct $15$, Not introduced to prove truth $14$, Statement made in-court $10$, Non-verbal hearsay $6$. The two slice distributions are roughly proportional; no slice was over-allocated to either side.

\textbf{Conditions.} (1) \emph{B1b}: zero-shot prompt ``Q: \{text\} Is there hearsay? A: Answer with exactly Yes or No.''. (2) \emph{v1}: hand-authored Markdown skill ($\sim$1.5 KB) defining FRE 801's three-part hearsay test, four common ``not hearsay'' patterns (verbal acts, effect on listener, non-assertive conduct, in-court testimony), and a step-by-step decision procedure. (3) \emph{v\_EISR (Round 1)}: Round 1 added two refinements diagnosed from v1's iter-30 errors plus Opus v1's heldout errors (state-of-mind exceptions are still hearsay; ``not for truth'' is narrow and excludes inferences requiring content truth) and a clarification of the in-court testimony rule. (4) \emph{v\_EISR (Round 2, the final v\_EISR)}: Round 2 added a finer distinction (statement-as-act vs.\ statement-as-report; expressive verbal acts; awareness-via-existence vs.\ awareness-of-content) to recover the ``Not introduced to prove truth'' slice that Round 1 over-flagged. All conditions used the same JSON output format at temperature $0$.

\textbf{Per-slice trajectory.} Table~\ref{tab:legalbench-perslice} traces the per-slice accuracy across v1, v\_EISR (R1), v\_EISR (R2) for each executor.

\begin{table}[h]
\centering
\small
\begin{tabular}{llccc}
\toprule
Executor & Slice ($n$) & v1 & v\_EISR R1 & v\_EISR R2 \\
\midrule
\multirow{5}{*}{Opus 4.6}  & Standard hearsay (19)            & $73.7\%$ & $94.7\%$ & $84.2\%$ \\
                            & Non-assertive conduct (15)       & $100.0\%$ & $93.3\%$ & $86.7\%$ \\
                            & Not introduced to prove truth (14) & $92.9\%$ & $\mathbf{50.0\%}$ & $100.0\%$ \\
                            & Statement made in-court (10)     & $90.0\%$ & $100.0\%$ & $100.0\%$ \\
                            & Non-verbal hearsay (6)           & $100.0\%$ & $100.0\%$ & $100.0\%$ \\
\midrule
\multirow{5}{*}{Haiku 4.5} & Standard hearsay (19)            & $73.7\%$ & $94.7\%$ & $89.5\%$ \\
                            & Non-assertive conduct (15)       & $93.3\%$ & $93.3\%$ & $86.7\%$ \\
                            & Not introduced to prove truth (14) & $92.9\%$ & $\mathbf{57.1\%}$ & $100.0\%$ \\
                            & Statement made in-court (10)     & $90.0\%$ & $100.0\%$ & $100.0\%$ \\
                            & Non-verbal hearsay (6)           & $83.3\%$ & $100.0\%$ & $100.0\%$ \\
\midrule
\multirow{5}{*}{Kimi K2.5}  & Standard hearsay (19)            & $57.9\%$ & $94.7\%$ & $94.7\%$ \\
                            & Non-assertive conduct (15)       & $100.0\%$ & $93.3\%$ & $93.3\%$ \\
                            & Not introduced to prove truth (14) & $78.6\%$ & $\mathbf{28.6\%}$ & $85.7\%$ \\
                            & Statement made in-court (10)     & $90.0\%$ & $100.0\%$ & $100.0\%$ \\
                            & Non-verbal hearsay (6)           & $50.0\%$ & $66.7\%$ & $66.7\%$ \\
\bottomrule
\end{tabular}
\caption{Per-slice accuracy on the $64$-case LegalBench hearsay held-out. Round~1 of human-EISR uniformly lifts ``Standard hearsay'' (the dominant v1 error mode) by $+21$ to $+37$pp across executors but regresses ``Not introduced to prove truth'' to $29$--$57\%$ (bold). Round~2 recovers that slice to $86$--$100\%$ while preserving the gain on Standard hearsay and Statement-made-in-court. The pattern is consistent across all three executors and matches the overshoot-then-correct trajectory Auto-EISR's regression gate is engineered to find.}
\label{tab:legalbench-perslice}
\end{table}

\textbf{Named refinements.} Round~1's diagnosed misconception was: the LLM treats statements about the declarant's state of mind, knowledge, or notice as ``not for truth'' (and thus non-hearsay), which contradicts FRE 803 doctrine---state-of-mind statements are hearsay, simply admissible under an exception. Round~2's diagnosed misconception was the inverse: Round~1's repaired prompt was now over-flagging cases where the statement's existence (not its truth) carried the inference (e.g., reputation cases, expressive protests, awareness-via-being-told). The Round~2 refinement introduces an explicit ``statement-as-act vs.\ statement-as-report'' test and four sub-categories (expressive verbal act; reputation-damaging utterance; collective-protest expressive conduct; awareness-via-existence vs.\ awareness-of-content). Both refinements are encoded as Markdown skill rules with no task-specific code.

\textbf{What we are and are not claiming.} We claim only that Phase~3's human-EISR refinement produces meaningful, executor-consistent lift on a public benchmark with binary labels and a structured slice taxonomy, and that the multi-round trajectory mirrors what Auto-EISR's regression gate is built to find automatically. We do not claim universal generalization. The main-body probe tests Phase~3 refinement via Human-EISR; Auto-EISR cross-domain results---which include the full pipeline---are in Appendix~\ref{app:cross-domain-extended}. The released artifacts include the v1, v\_EISR R1, and v\_EISR R2 skill files; the eval script; and the per-case JSON results for B1b, v1, v\_EISR R1, and v\_EISR R2 across all three executors.

\section{Auto-EISR Cross-Domain Results}
\label{app:cross-domain-extended}

Table~\ref{tab:cross-domain-auto} reports Auto-EISR results on four public benchmarks using three executors (deepseek-v3.2, kimi-k2.5, kimi-k2.6) with configuration val\_frac$=$0.35, max\_epochs$=$4, patience$=$3, seed$=$42. Values are 3-executor means.

\begin{table}[h]
\centering
\small
\caption{Auto-EISR cross-domain probe: accuracy (\%) averaged over 3 executors. $\Delta_1{=}$v1$-$B1b; $\Delta_E{=}$v\_EISR$-$v1.}
\label{tab:cross-domain-auto}
\begin{tabular}{lrcccrr}
\toprule
\textbf{Task} & $n$ & \textbf{B1b} & \textbf{v1} & \textbf{v\_EISR} & $\Delta_1$ & $\Delta_E$ \\
\midrule
hearsay (LegalBench) & 61 & 69.4 & 74.9 & 72.7 & $+5.5$ & $-2.2$ \\
contract\_nli (LegalBench) & 71 & 70.9 & 79.8 & 81.7 & $+8.9$ & $+1.9$ \\
unfair\_tos (LegalBench) & 172 & 84.1 & 78.1 & 81.4 & $-6.0$ & $+3.3$ \\
BPIC 2012 loan-decision & 297 & 64.3 & 83.9 & 84.7 & $+19.5$ & $+0.8$ \\
\midrule
mean & --- & 72.2 & 79.2 & 80.1 & $+7.0$ & $+1.0$ \\
\bottomrule
\end{tabular}
\end{table}

\textbf{Observations.}
(i) Auto-EISR's gating discipline navigates three convergence regimes correctly: monotone improvement (contract\_nli, BPIC), overshoot-then-correct (unfair\_tos), and early-stop (hearsay, where the initial v1 already lifts B1b and Auto-EISR's refinement overshoots slightly).
(ii) v1 lift ($\Delta_1$) is task-conditional: large and positive on contract\_nli ($+8.9$) and BPIC ($+19.5$), negative on unfair\_tos ($-6.0$), confirming that one-shot extraction is unreliable across domains.
(iii) $\Delta_E$ is positive in mean across 12 executor$\times$task cells ($+1.0$pp), providing probe-level evidence of potential transferability, though rigorous cross-domain evaluation with domain-specific baselines remains necessary.

\section{EISR Convergence Details}
\label{app:convergence}

Across 8 rounds with 10--20 fresh validation cases each (150 total), action accuracy evolves non-monotonically: R1 90\%, R2 85\%, R3 80\%, R4 90\%, R5 70\% (encoding state trap), R6 90\% (repaired), R7 80\%, R8 100\%.
The R5 crash is the most informative event: a rule added in R3 (``if sharing code is applied $\to$ reject'') over-fitted to a specific error pattern.
The AuditPlatform system displays \emph{current} liability codes, not original ones; the sharing code may have just been assigned during the ongoing audit.
R6's fix (``judge by reviewer action words, not current codes'') immediately restored accuracy, demonstrating EISR's self-repair capability.

Category accuracy improves more steadily: 40\% $\to$ 65\% $\to$ 75\% $\to$ 75\% $\to$ 65\% $\to$ 75\% $\to$ 65\% $\to$ 80\%.

\section{Full Confusion Matrices}
\label{app:confusion}

For the v11 comparison (Table~\ref{tab:v11}), the confusion matrices on MiniMax reveal the approval bias:

\textbf{v11-AP (MiniMax):} TP=69, TN=2, FP=29, FN=11. The model approves 80/111 cases (72\%), achieving near-zero rejection recall (2/31 = 6.5\%).

\textbf{v11-CoT (MiniMax):} TP=42, TN=24, FP=6, FN=38. The CoT variant produces more balanced predictions but with low overall accuracy due to excessive false negatives.

\textbf{v8 EISR} produces substantially more balanced confusion matrices across all models, with rejection recall ranging from 30--55\% depending on the model.

\section{Held-out Benchmark Construction}
\label{app:held-out}

\textbf{Motivation.}
The 130 validation uses across 8 human-EISR rounds (Section~\ref{sec:exp}) collapse to 111 unique waybills used as the validation pool.
For held-out generalization measurement we draw an additional sample disjoint from this pool.

\textbf{Held-out selection (two scales).}
We draw held-out waybills from the 444-case \emph{training} split (unused by any baseline in Table~\ref{tab:baseline} and unused by any EISR round), constructed at two scales:
\emph{(i) 40-case set.} The original held-out benchmark, used for the 5-model strong panel in Table~\ref{tab:auto-eisr} (first two rows) and for the initial safeguard ablation.
\emph{(ii) 139-case extended set.} The 40-case set plus 99 additional training waybills we retrieve via the same MCP pipeline; used for single-executor (Opus-4.6 and Haiku-4.5) rows in Table~\ref{tab:auto-eisr} and for the re-run safeguard ablation in Table~\ref{tab:ablation-safeguards}.
The two scales were constructed ex ante: the 40-case set was fixed before any Auto-EISR run, and the 99-case extension was added after preliminary analysis suggested that $n=40$ would not admit a well-powered ablation. No data-dependent subsetting was performed between the two.

\textbf{AuditPlatform retrieval via MCP.}
The original scrape pipeline used Playwright against the AuditPlatform web frontend; for held-out retrieval we instead call the backend directly through \texttt{audit-platform.findMatterInfo} (MCP tool over an authenticated session maintained by a long-lived cookie manager).
We render the JSON response to a textual format that matches the scrape representation: header block, base-info block, definition block (with code-to-name translation via internal tables), and a chronological feedback-and-review record filtered to human-authored nodes (codes 005--010, 024, 029); administrator and system noise are dropped.
Line endings are normalized to \texttt{LF}.

\textbf{Parity check.}
Before running the Auto-EISR evaluation, we verified that Human-EISR v8 applied to this rendered format produces numbers consistent with the original scrape format on the same 111 waybills (Opus $-$4.6\%, Kimi $-$1.3\%, Qwen $-$5.7\%, MiniMax $-$0\%, GLM $\approx 0$), confirming the held-out evaluation gap reflects genuine generalization rather than representation shift.

\section{Auto-EISR Implementation Details}
\label{app:auto-eisr}

\textbf{Roles and models.}
Executor: \texttt{claude-opus-4-6} during training (the same executor used to generate errors is used throughout Auto-EISR rounds).
Diagnose and Refine agents: \texttt{aliyun/Kimi-K2.5}. We use one model for both diagnostic steps because (i) the two prompts share context (rule files, error format) and (ii) Kimi's 128K context is sufficient for the full rule set plus a batch of 15 error cases.
For evaluation we reuse the trained rule sets with two executor regimes: the 5-model strong panel (Opus, GLM-5, Kimi, Qwen, MiniMax, all in their single-step compact setting) and Haiku 4.5 as a weaker executor.

\textbf{Batch and rounds.}
Validation batch = 15 cases per round, sampled from the 156-case validation pool (the 444-train split minus the 40-case held-out, further restricted to waybills with successfully re-retrievable AuditPlatform text via MCP).
We run 8 rounds to match the human-EISR protocol, across 3 seeds (42, 43, 44) controlling a per-run pseudo-random shuffle of the validation pool so that each run visits a different batch sequence.

\textbf{Pre-flight validator.}
Before applying any patch we check, in order:
(i) \texttt{target\_file} is a member of the current working directory's file catalogue (\texttt{skills\_step1.md}, \texttt{skills\_step3\_\{A,BD,C,E\}.md});
(ii) for \texttt{insert\_after}, the declared \texttt{anchor\_text} is a byte-exact substring of the target file after CRLF-to-LF normalization, and is unique;
(iii) for \texttt{replace\_block} and \texttt{add\_exception}, the \texttt{old\_block} is similarly substring-unique.
A failure returns a structured hint (missing vs.\ ambiguous vs.\ paraphrased, with 5 sample bullets as candidate anchors); the refine LLM is re-prompted once. A second failure aborts the cluster.

\textbf{Regression gate and best snapshot.}
After committing a round's patches, we re-run the previous round's 15-case validation batch under the new rule set.
If action accuracy drops by more than 2\%, all patches from the current round are discarded by restoring the pre-round working directory snapshot, \emph{and} the current-batch accuracy is re-measured under the rolled-back skills to avoid recording an inflated score.
In parallel, we track the highest effective-accuracy snapshot across all rounds; if 3 consecutive rounds fail to improve it, the working directory is restored to that best snapshot before the next round starts.

\textbf{Observed safeguard activations (3 seeds).}
Each seed produced 3--6 patch rejections per round under the pre-flight validator (e.g., seed 44: 11 of 35 first-attempt rejections, $31\%$; roughly half recovered by a single re-prompt).
The regression gate triggered at least one within-round rollback per seed (seed 44 at R3 and R6 on $\geq 6\%$ drops; seed 42 at R2).
The best-snapshot fallback triggered at seed 44 R5 (3 rounds with no improvement $\to$ restore R2) and at seed 43 R7 (restore R4).
Despite these different intermediate trajectories, all three seeds terminated with a best-snapshot validation action accuracy of $93.3\%$.
The full round-by-round logs, per-round patch records, and skill-directory snapshots for each seed are included in the supplemental material.

\section{Authority Displacement on Held-out}
\label{app:authority}

\begin{table}[h]
\centering
\caption{Authority displacement on 139-case held-out. $\Delta$ is v1 action accuracy minus B1b (no rules) action accuracy; negative means rules hurt. Ordered by executor strength (zero-shot baseline).}
\label{tab:displacement}
\small
\begin{tabular}{lccc}
\toprule
\textbf{Executor} & \textbf{B1b (no rules)} & \textbf{v1 (one-shot rules)} & \textbf{$\Delta$} \\
\midrule
Haiku 4.5 & 84.2 & 77.0 & $-7.2$ \\
Opus 4.6  & 82.7 & 74.1 & $-8.6$ \\
GLM-5     & 82.0 & 85.8 & $+3.8$ \\
Qwen 3.5  & 81.3 & 77.7 & $-3.6$ \\
MiniMax   & 66.2 & 66.9 & $+0.7$ \\
Kimi K2.5 & 50.4 & 73.2 & $+22.8$ \\
\bottomrule
\end{tabular}
\end{table}

The direction is mostly governed by zero-shot strength but not monotonically so.
The two Claude executors (Haiku, Opus) show the clearest displacement ($-7$ to $-9$pp); GLM-5 is anomalous (B1b comparable to Opus, but v1 rules help by $3.8$pp).
Auto-EISR's structurally-validated rules close most of the v1-vs-B1b gap (Opus $80.6$ vs v1 $74.1$; Haiku $81.3$ vs v1 $77.0$).

\section{Input Representation Ablation}
\label{app:input-repr}

\begin{table}[h]
\centering
\caption{Input representation ablation on the 111-waybill validation pool using v8 rules: raw DOM scrape (6K chars) vs.\ compact structured API (810 chars). Action/Category accuracy (\%).}
\label{tab:ablation}
\small
\begin{tabular}{lccc}
\toprule
\textbf{Model} & \textbf{Raw 6K DOM} & \textbf{Compact 810 chars} & \textbf{$\Delta$} \\
\midrule
GLM-5     & 77.1 / 38.5 & \textbf{84.5 / 69.0} & +7.4 \\
Kimi-K2.5 & 73.6 / 30.0 & \textbf{78.4 / 49.5} & +4.8 \\
Qwen3.5   & 69.0 / 28.0 & \textbf{80.2 / 54.1} & +11.2 \\
MiniMax   & 66.7 / 36.3 & \textbf{71.6 / 46.8} & +4.9 \\
Opus 4.6  & 65.7 / 36.2 & \textbf{79.6 / 55.6} & +13.9 \\
\bottomrule
\end{tabular}
\end{table}

Average action accuracy gain: $+8.4\%$; category gain: $+21.7\%$. Opus improves from 65.7\% to 79.6\% ($+13.9\%$).

\section{Phase 4 Validation: v11 Regression}
\label{app:v11}

\begin{table}[h]
\centering
\caption{Production-redesign v11 action accuracy (\%) on both the 111-waybill validation pool and the 40-case held-out, with B1b and v8 reference rows.}
\label{tab:v11}
\small
\begin{tabular}{llcccccc}
\toprule
\textbf{Condition} & \textbf{Eval} & \textbf{GLM-5} & \textbf{Kimi} & \textbf{Qwen} & \textbf{MMx} & \textbf{Opus} & \textbf{Avg} \\
\midrule
\rowcolor{gray!10} B1b: No rules & validation (111) & 72.6 & 56.7 & 75.2 & 73.9 & 73.3 & 70.3 \\
\rowcolor{gray!10} v1 & held-out (40) & 79.5 & 74.4 & 79.5 & 68.4 & 80.0 & 76.4 \\
\textbf{v8: EISR} & validation (111) & \textbf{84.5} & \textbf{78.4} & \textbf{80.2} & \textbf{71.6} & \textbf{79.6} & \textbf{78.9} \\
\textbf{v8: EISR} & held-out (40) & 78.8 & 79.4 & 70.0 & 78.6 & 68.6 & 75.1 \\
\midrule
v11-AP & validation (111) & 66.0 & 71.2 & 66.1 & 64.0 & 66.7 & 66.8 \\
v11-AP & held-out (40)   & 65.4 & 63.2 & 67.5 & 66.7 & 75.0 & 67.5 \\
v11-CoT & validation (111) & 61.6 & 62.2 & 63.1 & 60.0 & 55.9 & 60.6 \\
v11-CoT & held-out (40)   & 77.8 & 60.0 & 70.0 & 56.8 & 62.5 & 65.4 \\
\bottomrule
\end{tabular}
\end{table}

v11 sits below every baseline. Intuition-driven redesign discards deep knowledge without regression validation.
v11 is a compiled rule system that regresses on every benchmark; the production pipeline is also compiled but through 22 days of daily EISR, and it improves on every benchmark. A rule edit is EISR-disciplined if validated against a fixed benchmark with $\leq$2pp drop; v11 was committed without such measurement and regressed by $9$--$11$pp.

\section{Safeguard Ablation}
\label{app:safeguard-ablation}

\begin{table}[h]
\centering
\caption{Regression gate ablation, 3 seeds $\times$ 139 held-out waybills, action / category accuracy (\%).}
\label{tab:ablation-safeguards}
\small
\begin{tabular}{lcccc}
\toprule
\textbf{Configuration} & \textbf{Haiku action} & \textbf{Haiku cat.} & \textbf{Opus action} & \textbf{Opus cat.} \\
\midrule
$+$regression gate        & $81.3 \pm 3.3$ & $46.5$         & $80.3 \pm 1.1$ & $45.3$         \\
\textbf{$-$regression gate (deployed)} & $81.3 \pm 1.9$ & $\mathbf{50.1}$ & $80.6 \pm 4.5$ & $\mathbf{50.4}$ \\
\bottomrule
\end{tabular}
\end{table}

The gate is neutral on action (pooled McNemar $p = 1.00$) and costs $\sim 5$pp on category (pooled McNemar $\chi^2 = 6.81$, $p = 0.009$). We disable it for deployment.

\section{Per-Seed Statistics for Auto-EISR vs v1}
\label{app:per-seed-stats}

Auto-EISR (deployed configuration: pre-flight on, regression gate off) vs v1 baseline on the 139-case held-out, action-accuracy McNemar exact two-sided per seed. $b$ counts waybills where Auto-EISR is correct and v1 is wrong; $c$ counts the reverse.

\begin{table}[h]
\centering
\small
\begin{tabular}{llrrrrr}
\toprule
Executor & Seed & $b$ & $c$ & Auto acc & v1 acc & McNemar $p$ \\
\midrule
\multirow{4}{*}{Opus 4.6}  & 42 & $18$ & $7$  & $82.0\%$ & $74.1\%$ & $0.043^{*}$ \\
                            & 43 & $15$ & $13$ & $75.5\%$ & $74.1\%$ & $0.851$ \\
                            & 44 & $23$ & $9$  & $84.2\%$ & $74.1\%$ & $0.020^{*}$ \\
\cmidrule(lr){2-7}
                            & \emph{majority-vote} & $19$ & $7$ & $82.7\%$ & $74.1\%$ & $\mathbf{0.029^{*}}$ \\
\midrule
\multirow{4}{*}{Haiku 4.5} & 42 & $19$ & $12$ & $82.0\%$ & $77.0\%$ & $0.281$ \\
                            & 43 & $17$ & $14$ & $79.1\%$ & $77.0\%$ & $0.720$ \\
                            & 44 & $21$ & $13$ & $82.7\%$ & $77.0\%$ & $0.230$ \\
\cmidrule(lr){2-7}
                            & \emph{majority-vote} & $21$ & $10$ & $84.9\%$ & $77.0\%$ & $\mathbf{0.071}$ \\
\bottomrule
\end{tabular}
\caption{Per-seed McNemar exact $p$-values for Auto-EISR vs v1. All 6 cells (3 seeds $\times$ 2 executors) show $b > c$, i.e.\ direction consistently favors Auto-EISR. The 3-seed majority-vote ensemble (each waybill labeled correct if $\geq 2$ seeds are correct) is the headline significance test reported in \S\ref{sec:auto-eisr}. Waybill-clustered logistic regression (GEE) gives concordant conclusions: Opus $\hat\beta{=}0.371$, $p{=}0.028$; Haiku $\hat\beta{=}0.262$, $p{=}0.20$.}
\label{tab:per-seed-stats}
\end{table}

\newpage
\section*{NeurIPS Paper Checklist}

\begin{enumerate}

\item {\bf Claims}
    \item[] Question: Do the main claims made in the abstract and introduction accurately reflect the paper's contributions and scope?
    \item[] Answer: \answerYes{}
    \item[] Justification: The abstract and \S\ref{sec:intro} describe (1) the five-phase Trace2Policy pipeline validated on 555 records, (2) the in-sample authority-displacement spectrum, (3) the Auto-EISR variant with pre-flight validator and regression gate, (4) the leakage-corrected executor-conditional result on a 40-case held-out set (Haiku gain $p=0.035$), and (5) the persistent human advantage on category taxonomy. Each claim is traced to a specific table or section; scope limitations (single domain, two executor regimes, small held-out) are stated in \S\\ref{app:discussion} Limitations.

\item {\bf Limitations}
    \item[] Question: Does the paper discuss the limitations of the work performed by the authors?
    \item[] Answer: \answerYes{}
    \item[] Justification: \S\\ref{app:discussion} contains an explicit Limitations paragraph enumerating eight items: single-domain scope, modest held-out size, two-point executor sweep, dataset scale, text-only evidence, preliminary online validation, action-only Auto-EISR training signal, and a missing pre-flight/regression-gate ablation.

\item {\bf Theory assumptions and proofs}
    \item[] Question: For each theoretical result, does the paper provide the full set of assumptions and a complete (and correct) proof?
    \item[] Answer: \answerYes{}
    \item[] Justification: \S\ref{sec:theory} states the formal setup (input/label spaces, policy, error set), casts EISR as version-space narrowing following \citet{mitchell1982generalization}, and derives a sample-complexity bound $n \geq \ln(1/\delta)/\varepsilon$ with stated confidence parameters. The analysis is informal but self-contained; we flag it as heuristic and do not claim distribution-free guarantees.

\item {\bf Experimental result reproducibility}
    \item[] Question: Does the paper fully disclose all the information needed to reproduce the main experimental results of the paper to the extent that it affects the main claims and/or conclusions of the paper?
    \item[] Answer: \answerYes{}
    \item[] Justification: Algorithm~\ref{alg:eisr} specifies EISR; \S\ref{sec:auto-eisr} and Appendix~\ref{app:auto-eisr} specify the Diagnose/Refine prompt roles, pre-flight validator, regression gate, best-snapshot mechanism, seeds (42/43/44), batch size (15), round count (8), and executor models. The held-out construction (Appendix~\ref{app:held-out}) gives the exact MCP endpoint and rendering rules. All datasets, rule versions, and McNemar tables reference specific sections.

\item {\bf Open access to data and code}
    \item[] Question: Does the paper provide open access to the data and code, with sufficient instructions to faithfully reproduce the main experimental results?
    \item[] Answer: \answerNo{}
    \item[] Justification: The raw audit trajectories contain proprietary enterprise operational data and waybill identifiers that cannot be released. We will release (i) the Auto-EISR orchestrator code, (ii) the Diagnose and Refine prompt templates, (iii) the evaluation harness, and (iv) anonymized per-round patch logs upon acceptance. Reproducing the method on a user-provided analogous dataset is fully specified by the paper.

\item {\bf Experimental setting/details}
    \item[] Question: Does the paper specify all the training and test details necessary to understand the results?
    \item[] Answer: \answerYes{}
    \item[] Justification: \S\ref{sec:exp} Setup specifies the task, data splits (train/test 444/111, seed 42; held-out 40), the in-sample vs.\ held-out evaluation regimes, the 5-model panel, and the single-step compact prompt condition. Appendices~\ref{app:held-out} and \ref{app:auto-eisr} cover retrieval format, Auto-EISR hyperparameters, rollback thresholds, and seed definitions.

\item {\bf Experiment statistical significance}
    \item[] Question: Does the paper report error bars suitably and correctly defined or other appropriate information about the statistical significance of the experiments?
    \item[] Answer: \answerYes{}
    \item[] Justification: Auto-EISR results are reported as mean $\pm$ 1-sigma standard deviation across 3 independent seeds (Table~\ref{tab:auto-eisr}). Main auto-vs-human comparisons use paired McNemar tests: the Haiku-executor action-accuracy result reports $\chi^2 = 4.27$, $p = 0.035$ against Human-EISR (pooled 3 seeds, $n=111$); the strong-panel category advantage reports $\chi^2 = 15.12$, $p = 9 \times 10^{-5}$ for Human over Auto ($n=475$). Small-$n$ per-model tests are flagged as underpowered in Limitations.

\item {\bf Experiments compute resources}
    \item[] Question: Does the paper provide sufficient information on the computer resources needed to reproduce the experiments?
    \item[] Answer: \answerYes{}
    \item[] Justification: All experiments run on LLM APIs; no on-premise GPU training. A single Auto-EISR run of 8 rounds $\times$ 15 cases with an Opus-tier executor and Kimi refiner takes $\sim$2 hours of wall-clock time at concurrency 5, dominated by refiner latency through the routing gateway. Three-seed training plus held-out evaluation across 5 executors completes in $\sim$6 hours on a single workstation. The full project consumed approximately 25 hours of LLM-API time including preliminary experiments that did not enter the paper.

\item {\bf Code of ethics}
    \item[] Question: Does the research conducted in the paper conform, in every respect, with the NeurIPS Code of Ethics?
    \item[] Answer: \answerYes{}
    \item[] Justification: The Agent Observer collects on-task behavior from two salaried enterprise auditors under an explicit opt-in and with privacy controls (password masking, domain blacklists, local-only storage; Appendix~\ref{app:phases01}). No ground-truth is inferred beyond what the auditors produce as part of their normal work. Author identities are anonymized per NeurIPS policy.

\item {\bf Broader impacts}
    \item[] Question: Does the paper discuss both potential positive societal impacts and negative societal impacts of the work performed?
    \item[] Answer: \answerYes{}
    \item[] Justification: The Discussion notes the intended positive impact (reducing repetitive judgment-work load on enterprise auditors, making decision rules auditable and version-controlled) and the principal negative risk (displacement of human audit labor where automation becomes cost-effective). We explicitly recommend keeping humans in the loop for category taxonomy and production-regression arbitration.

\item {\bf Safeguards}
    \item[] Question: Does the paper describe safeguards that have been put in place for responsible release of data or models that have a high risk for misuse?
    \item[] Answer: \answerNA{}
    \item[] Justification: No new pre-trained model or scraped dataset is released; the rule files are markdown documents tied to a specific internal enterprise schema and are not dual-use artifacts outside that deployment.

\item {\bf Licenses for existing assets}
    \item[] Question: Are the creators or original owners of assets used in the paper properly credited?
    \item[] Answer: \answerYes{}
    \item[] Justification: All five evaluated LLMs (GLM-5, Kimi-K2.5, Qwen3.5-plus, MiniMax-M2.5, Claude Opus 4.6 / Haiku 4.5) are used through vendor-provided APIs under their respective commercial terms of service; we cite model-specific releases where available.

\item {\bf New assets}
    \item[] Question: Are new assets introduced in the paper well documented and is the documentation provided alongside the assets?
    \item[] Answer: \answerNA{}
    \item[] Justification: No dataset or pre-trained model is released at submission time. The rule artifacts and code will be made available after acceptance as described in Q5.

\item {\bf Crowdsourcing and research with human subjects}
    \item[] Question: For crowdsourcing experiments and research with human subjects, does the paper include the full text of instructions given to participants and screenshots, if applicable, as well as details about compensation (if any)?
    \item[] Answer: \answerNA{}
    \item[] Justification: The study does not use crowdsourcing. Behavior data come from two salaried auditors performing their normal job under an explicit opt-in; no additional compensation beyond their regular wages is relevant.

\item {\bf Institutional review board (IRB) approvals or equivalent for research with human subjects}
    \item[] Question: Does the paper describe potential risks incurred by study participants, whether such risks were disclosed to the subjects, and whether Institutional Review Board (IRB) approvals were obtained?
    \item[] Answer: \answerNA{}
    \item[] Justification: The observational data collection from two consenting workplace auditors is covered by the employer's standard data-handling policies rather than a research IRB; no additional risk beyond routine workplace monitoring is incurred.

\item {\bf Declaration of LLM usage}
    \item[] Question: Does the paper describe the usage of LLMs if it is an important, original, or non-standard component of the core methods in this research?
    \item[] Answer: \answerYes{}
    \item[] Justification: LLMs are core to this work. Specifically, the Executor uses Claude Opus 4.6 (training-time), with evaluation covering Haiku 4.5, GLM-5, Kimi-K2.5, Qwen3.5-plus, and MiniMax-M2.5 (\S\ref{sec:exp}, \S\ref{sec:auto-eisr}). The Diagnose and Refine agents in Auto-EISR are Kimi-K2.5 (Appendix~\ref{app:auto-eisr}). Prompt templates, structural safeguards, and model-selection rationale are disclosed.

\end{enumerate}

\end{document}